\RequirePackage{letltxmacro}
\LetLtxMacro{\LaTeXtextbf}{\textbf}
\documentclass{ieeeaccess}
\LetLtxMacro{\textbf}{\LaTeXtextbf}

\makeatletter \def\input@path{ {inputs/} } \makeatother

\usepackage[compress]{cite}
\usepackage{amsmath}
\usepackage{amssymb}
\usepackage{amsfonts}
\usepackage{graphicx}
\usepackage{textcomp}

\let\oldyear\year
\def\year{%
  \ifdefined\pdfprimitive
    \expandafter\pdfprimitive
  \else
    \expandafter\primitive
  \fi
  \year}
\usepackage{expl3}
\let\year\oldyear

\usepackage[hyphens,spaces,obeyspaces]{url} 
\usepackage[
  bookmarks=true,
  hidelinks=true,
  ]{hyperref}
\usepackage{booktabs} 
\usepackage[binary-units=true]{siunitx}[=v2]  
\usepackage{dsfont}   
\usepackage[linesnumbered,algoruled,boxed,lined,noend]{algorithm2e}
\usepackage{algpseudocode}
\usepackage{color}
\usepackage{multirow}
\usepackage{paralist}
\usepackage{mathrsfs}   
\usepackage{bm}         
\usepackage{xspace}     

\DeclareSIUnit[number-unit-product = ]\percent{\%}

\usepackage{etoolbox}
\robustify\bfseries
\def\B{\bfseries \color{darkgreen}}

\renewcommand{\vec}[1]{\bm{#1}}

\newcommand{\Sx}{S_{\!\times\!}}
\renewcommand{\emptyset}{\{\}}
\newcommand{\etal}{\textit{et al.}\xspace}

\newcommand{\predicate}[1]{\nobreak{\textnormal{#1}}\xspace}
\newcommand{\collisionFree}{\predicate{CollisionFree}}
\newcommand{\selfCollisionFree}{\predicate{SelfCollisionFree}}
\newcommand{\voxelIdx}{\predicate{VoxelIndex}}

\newcommand{\algname}[1]{\predicate{#1}}
\newcommand{\NormalIK}{\algname{NormalIK}}
\newcommand{\RoadmapIK}{\algname{RoadmapIK}}
\newcommand{\VoxelizeFreeEdge}{\algname{VoxelizeFreeEdge}}
\newcommand{\VoxelizeFreeEdgeRecurse}{\algname{VoxFreeEdgeRecurse}}
\newcommand{\VoxelizeBackbone}{\predicate{VoxelizeBackbone}}

\newcommand{\qbegin}{\vec{q}_{\text{a}}}
\newcommand{\qend}{\vec{q}_{\text{b}}}
\newcommand{\qmid}{\vec{q}_{\text{mid}}}
\newcommand{\qreached}{\vec{q}_{\text{reached}}}

\DeclareMathOperator*{\argmin}{arg\,min}

\definecolor{purple}{RGB}{210, 0, 210}
\definecolor{darkred}{RGB}{150, 0, 0}
\definecolor{greenishblue}{RGB}{37, 132, 172}
\definecolor{blueishgreen}{RGB}{0, 172, 72}
\definecolor{orange}{RGB}{255, 150, 0}
\definecolor{darkgreen}{RGB}{0, 100, 0}
\definecolor{darkblue}{RGB}{30, 30, 180}
\definecolor{darkgrey}{RGB}{130, 130, 130}

\newcommand{\authfig}[1]{%
  {\includegraphics[width=1in,height=1.25in,clip,keepaspectratio]{#1}}%
}

\author{%
  \uppercase{Michael Bentley\authorrefmark{1},}
  \uppercase{Caleb Rucker\authorrefmark{2},} \IEEEmembership{Senior Member, IEEE},
  \uppercase{and Alan Kuntz\authorrefmark{1},} \IEEEmembership{Member, IEEE}%
}
\address[1]{%
  Robotics Center and School of Computing,
  University of Utah, Salt Lake City, UT 84112 USA
  (e-mail: michael.bentley@utah.edu; alan.kuntz@utah.edu)%
}
\address[2]{%
  University of Tennessee, Knoxville, TN 37996 USA
  (e-mail: caleb.rucker@utk.edu)%
}

\corresp{Corresponding author: Michael Bentley (e-mail: michael.bentley@utah.edu).}

\graphicspath{{images/}}

\tfootnote{
  The work of Michael Bentley and Alan Kuntz was supported in part by the NSF under Grant \#2133027.
  The work of Caleb Rucker was supported in part by the NSF under Grant IIS-1652588.
}

\title{
  Interactive-Rate Supervisory Control for Arbitrarily-Routed Multi-Tendon Robots via Motion Planning
  }
\history{Received 24 May 2022, accepted 16 July 2022, date of publication 28 July 2022, date of current version 5 August 2022.}
\doi{\href{https://doi.org/10.1109/ACCESS.2022.3194515}{10.1109/ACCESS.2022.3194515}}
\vol{10}
\year{2022}
\pubid{
  This work is licensed under a Creative Commons Attribution 4.0 License.
  For more information, see \url{https://creativecommons.org/licenses/by/4.0/}
}

\begin{document}

\begin{abstract}
  Tendon-driven robots, where one or more tendons under tension bend and manipulate a flexible backbone, can improve minimally invasive surgeries involving difficult-to-reach regions in the human body.
  Planning motions safely within constrained anatomical environments requires accuracy and efficiency in shape estimation and collision checking.
  Tendon robots that employ arbitrarily-routed tendons can achieve complex and interesting shapes, enabling them to travel to difficult-to-reach anatomical regions.
  Arbitrarily-routed tendon-driven robots have unintuitive nonlinear kinematics.
  Therefore, we envision clinicians leveraging an assistive interactive-rate motion planner to automatically generate collision-free trajectories to clinician-specified destinations during minimally-invasive surgical procedures.
  Standard motion-planning techniques cannot achieve interactive-rate motion planning with the current expensive tendon robot kinematic models.
  In this work, we present a 3-phase motion-planning system for arbitrarily-routed tendon-driven robots with a Precompute phase, a Load phase, and a Supervisory Control phase.
  Our system achieves an interactive rate by developing a fast kinematic model (over \num{1 000} times faster than current models), a fast voxel collision method (\num{27.6} times faster than standard methods), and leveraging a precomputed roadmap of the entire robot workspace with pre-voxelized vertices and edges.
  In simulated experiments, we show that our motion-planning method achieves high tip-position accuracy and generates plans at \SI{14.8}{\Hz} on average in a segmented collapsed lung pleural space anatomical environment.
  Our results show that our method is \num{17 700} times faster than popular off-the-shelf motion planning algorithms with standard FK and collision detection approaches.
  Our open-source code is available online.
\end{abstract}


\titlepgskip=-15pt

\maketitle

\section{Introduction}
\label{sec:intro}
%
%
%
%

%
%
\PARstart{C}{ontinuum} robots, flexible robots capable of taking curvilinear shapes, have the potential to reduce the invasiveness of numerous surgical applications~\cite{Burgner-Kahrs2015_TRO,Dupont2022_ProcIEEE}.
Their potential in surgical applications comes from their ability to curve around anatomical structures to access sites that traditional surgical tools cannot reach while maintaining compliance.
One promising type of continuum robot is the tendon-driven robot~\cite{Kato2015_TM,Kutzer2011_ICRA,Nguyen2015_IROS}.
Tendon-driven robots have a flexible backbone with cables or tendons connected at some distance away from the backbone, usually by disks rigidly attached to the backbone.
Tension on the tendons actuates the robot backbone, bending it according to each tendon's routing and tension.
Traditionally, the tendons are routed straight along the length of the robot's shaft, producing a constant curvature shape when pulled.
However, tendons with more complex routings can bend the robot out-of-plane and with non-constant curvature~\cite{Oliver-Butler2019_TRO,Rucker2011_TRO,Huang2021_ISMR,Starke2017_IROS}.
With multiple complex-routed tendons, various expressive shapes and motions are possible, opening the door for safer operation in complex patient anatomy~\cite{Starke2017_IROS}.

\begin{figure*}
  \centering
  \includegraphics[width=\textwidth]{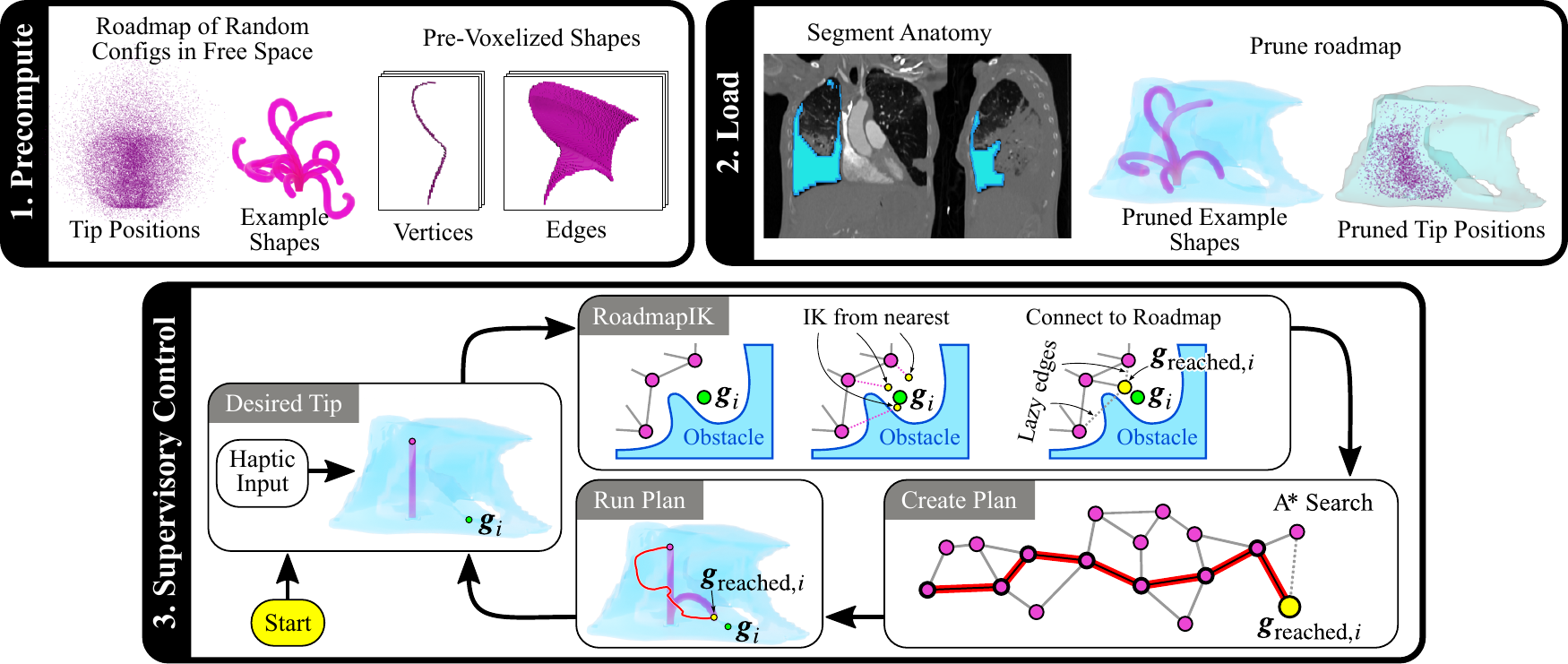}
  \caption{
    An overview of the three phases of our proposed method for robot-assisted interactive-rate motion planning of a tendon-driven robot.
    \textbf{1. Precompute:} 
    This phase involves precomputing the robot shape of many random configurations, connecting them together into a graph called a roadmap, and voxelizing their centerline and motions between connected configurations. We show the tip positions of \num{30 000} random robot configurations, the full shape of ten example configurations from the set of \num{30 000}, an example voxelized robot shape along the centerline, and an example of a voxelized swept volume of the robot centerline as the robot moves between two robot configurations. The full backbone voxelization is stored for all robot configurations and the edge voxelization for all edges. 
    \textbf{2. Load:} In this phase, we prune away in-collision portions of the precomputed roadmap when presented with a specific anatomical environment.
    We show two cross-section images for an example Computed Tomography (CT) scan where we segmented a pleural space of a collapsed lung in which the robot will be operating (blue).
    For the ten example robot shapes shown on the left in the Precompute phase, we show the remaining five collision-free shapes and the remaining tip positions of the many more not-shown collision-free configurations from the example tip positions shown in the Precompute Phase.
    \textbf{3. Supervisory Control:} This phase generates motion plans for specified tip positions within the anatomical region.
    The cycle starts by obtaining a desired tip position $\vec{g}_i$, as provided by a clinician, for example through a haptic input device.
    Then, the \RoadmapIK algorithm generates an inverse-kinematic (IK) solution $\vec{g}_{\text{reached},i}$ close to $\vec{g}_i$ from nearby configurations on the roadmap, such that it has one known collision-free edge connecting it to the roadmap and has other lazily-evaluated edges to other nearby roadmap vertices.
    Next, the system creates a plan over our roadmap with the A$\!^*$ search algorithm to $\vec{g}_{\text{reached},i}$, executes the plan on the robot, and then waits for the next desired tip position, continuing the cycle until the procedure has ended.
  }
  \label{fig:fig1}
\end{figure*}

\begin{figure}
  \centering
  \includegraphics[width=.65\columnwidth]{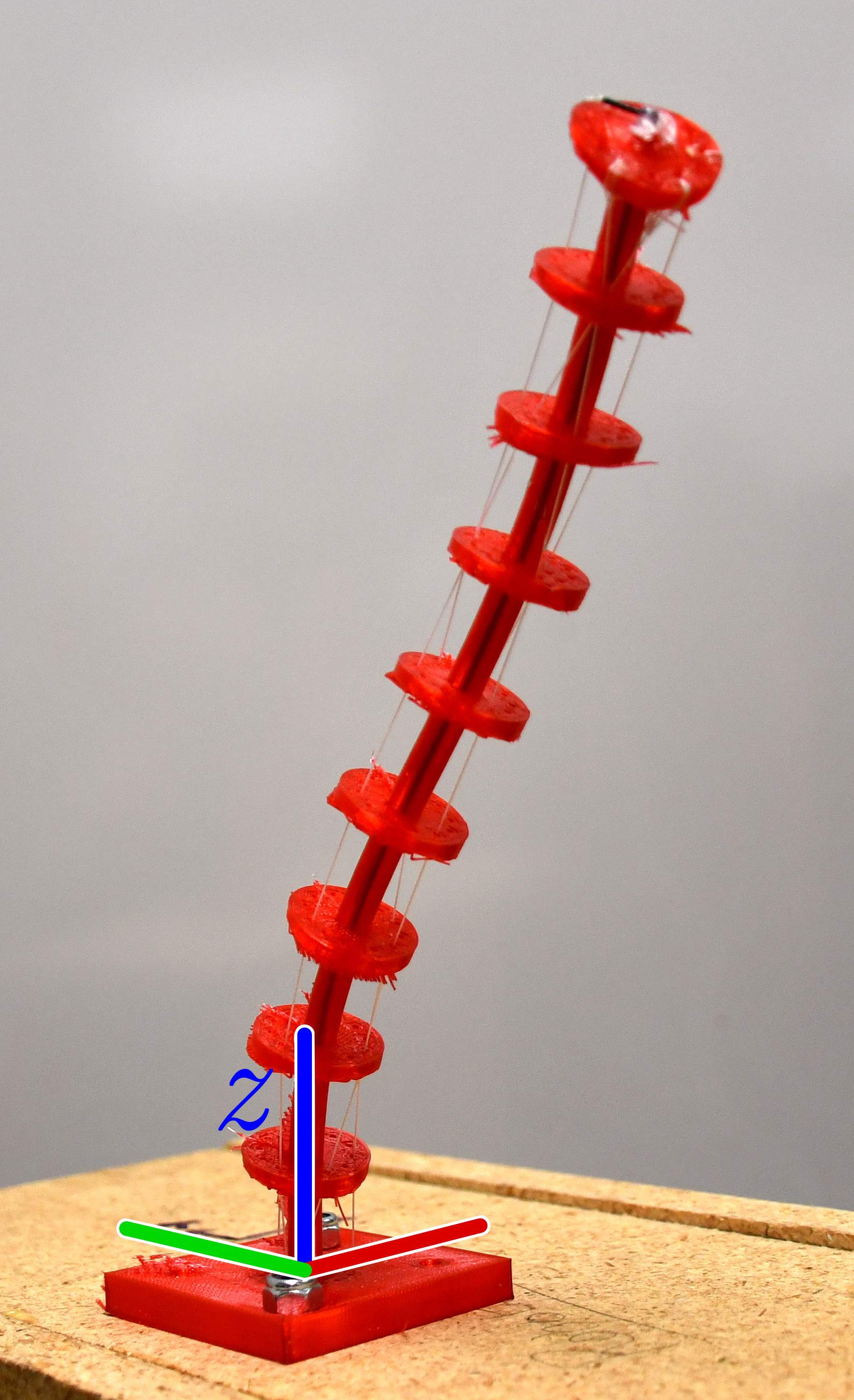}
  \caption{
    Our physical tendon robot actuated by multiple tendons.
    The backbone, disks, and base mounting plate are 3D printed in flexible TPU as a single piece.
    We actuate the robot in this picture using three of the four tendons into an interesting shape only achievable with non-straight tendon routings.
    The robot's base frame is shown with the $z$-axis explicitly labeled.
  }
  \label{fig:tendon-robot}
\end{figure}

However, manual control of these devices is infeasible due to the complicated relationship between the tendon tensions/displacements and the robot's shape.
Robot-assisted supervisory control enables a computer to control the robot's complex kinematics while the user specifies end-effector positions or poses.
Such supervisory control enables intuitive human operation in traditional surgical robots and continuum robots alike~\cite{Lynch2017_book}.
However, during supervisory control, unintended collisions of the robot's shaft with patient anatomy may cause damage to the patient, inaccurate control, and potentially result in adverse outcomes.
Avoiding such collisions is challenging with tendon-driven robots in traditional teleoperation methods due to the complex way the robot's end-effector motion relates to the shape change of the robot's body proximal to the end effector---small motions at the robot's tip may require extensive and unintuitive motions along the robot's body.

Leveraging motion planning in the supervisory control of continuum robots may mitigate these concerns by ensuring that collisions with the patient's anatomy are avoided during robot motions while providing the user with intuitive end-effector control~\cite{Torres2014_ICRA,Torres2015_ICRA}.
The main challenges preventing us from applying traditional motion-planning algorithms to our problem are the complex and slow kinematic model of tendon-driven continuum robots and the difficulty of reaching the desired tip position without a method for goal biasing or an efficient inverse kinematics (IK) solver.

%
%

This work overcomes these challenges and provides a fast and safe motion-planning system for a tendon-driven robot that interactively plans collision-free paths to user-specified robot goal tip positions.
Our approach is separated into three phases, as seen in Fig.~\ref{fig:fig1}:
\begin{enumerate}
  \item
    a \textit{Precompute phase} agnostic to a specific anatomical environment,
  \item
    a \textit{Load phase}, loading and integrating the precomputed pieces into a specific anatomical environment, and
  \item
    an interactive \textit{Supervisory-Control phase}, where the motion planner generates collision-free paths from user-specified goal-tip positions.
\end{enumerate}
Fig.~\ref{fig:tendon-robot} shows an example tendon-driven robot we 3D printed, demonstrating an interesting shape achievable with complex-routed tendons.

The \textbf{Precompute phase} randomly samples robot configurations and connects them to make a dense roadmap graph.
We precompute and store voxelized robot shapes and swept-volume motions with the roadmap, enabling fast collision detection in subsequent phases and avoiding many kinematic shape computations at planning time.
This phase only needs to be performed once for the robot as it is agnostic to any specific anatomical environment.
When next presented with a specific anatomical environment in which the robot will be operating, e.g., segmented from medical imaging of a patient prior to a medical procedure (see Fig.~\ref{fig:fig1}), we refine the precomputed roadmap during the load phase.

The \textbf{Load phase} loads the precomputed roadmap and, using the vertex and edge voxelizations, performs fast collision detection between the environment segmentation and the roadmap, pruning away roadmap vertices and edges that are in collision with the environment.
This pruning results in a dense, collision-free roadmap representing the free space in which the robot operates during the procedure via the supervisory-control phase.

During the \textbf{Supervisory-Control phase}, the clinician specifies desired tip positions for the robot interactively.
For each specified tip position, our method generates robot configurations that place the tip at or near the specified point, connects the closest of these generated configurations to the pruned roadmap, and plans a collision-free motion to this connected goal configuration.
Then the motion plan executes, moving the robot's tip to the clinician-specified point.
This process repeats as the clinician specifies a new goal tip position starting where the last plan ended.

%
%

This paper provides four main contributions that enable our three-phase motion-planning system.
\begin{enumerate}
  \item
    We present a faster model for computing forward-kinematics (FK) of an arbitrarily-routed tendon-driven robot in the case of no external forces and torques.
  \item
    We present a voxel-to-voxel collision detection method that provides significantly faster collision checking than traditional mesh-based methods while leveraging the original environment representation of voxels from typical 3D medical imaging segmentation.
  \item
    We present a strategy for motion validation and swept-volume voxelization that discretizes dynamically based on voxel dimensions.
  \item
    We develop an inverse-kinematics (IK) algorithm that connects multiple nearest neighbors in the precomputed roadmap to specified desired robot tip positions, enabling effective interactive supervisory control.
\end{enumerate}
Our system achieves interactive-rate robot-assisted supervisory control of tendon-driven robots through the above contributions.
The code, documentation, and additional supplemental information is available at \url{https://sites.google.com/gcloud.utah.edu/armlab-tendon-planning}.

%
%
We first show that our improved FK solver is more than \num{1 000} times faster when compared against a standard FK implementation.
We next demonstrate and evaluate our methods' efficacy by simulating the tendon-driven robot moving in a patient's pleural cavity, the free space between the patient's chest wall and collapsed lung, both near the surface of anatomy and throughout the pleural volume.
We compare our method against standard motion-planning algorithms and perform an ablation study to evaluate each aspect of our method in computational performance and accuracy in tip position.
Our method demonstrates significant improvements over competing motion-planning algorithms in both tip position accuracy and in planning time.
We achieve motion planning at an interactive rate of \SI{14.8}{\Hz} on average, more than \num{33} times faster than the best evaluated competing algorithm when allowing competing algorithms to use our fast FK and collision methods, and more than \num{17 700} times faster than competing planners without the aid of our other contributions.

This work enables safe, interactive-rate tip-position supervisory control of a tendon-driven robot via motion planning, opening the door for the potential use of these promising continuum medical devices to reduce surgical procedures' invasiveness and improve patient outcomes.

\section{Related Work}
\label{sec:relatedwork}

%
%
Due to their ability to curve around anatomical structures, continuum robots are particularly promising for medical applications~\cite{Burgner-Kahrs2015_TRO,Peters2018_SE,Dupont2022_ProcIEEE}.
Tendon-driven robots in particular have potential application in many surgical domains including endoscopic surgery~\cite{Kato2015_TM,Kutzer2011_ICRA,Nguyen2015_IROS,Huang2021_ISMR}.
In this work, we consider the application of robot-assisted supervisory control by a clinician within an anatomical cavity, such as the one shown in Fig.~\ref{fig:fig1}.

%
%
Modeling for continuum robots is complex and typically computationally expensive.
Many models for tendon-driven robots consider multiple separate tendon-actuated segments using simple straight tendon routings to simplify the kinematic model to a sequence of circular arcs~\cite{Zhang2021_JAIT,Neppalli2009_AR,Xu2009_JMR,Lai2022_RAL}.
Although this is very efficient to compute, this approach restricts the types of shapes achievable by the robot.
Static and dynamic models for tendon-driven robots with arbitrary tendon routings and external forces have been developed using Cosserat Rod and Cosserat String models~\cite{Rucker2011_TRO,Oliver-Butler2019_TRO}.
However, these models are expensive to compute as they rely on solving a boundary-value problem (BVP) on a linear set of differential equations.
More recent work uses optimization and geometric considerations for kinematic modeling with one arbitrarily-routed tendon~\cite{Ashwin2019_AMMS,Mahapatra2022_AAMMS}.
However, this geometric approach can only handle one tendon, which limits its applicability considerably.
In this work, we adapt the static model in~\cite{Rucker2011_TRO}, which allows for many arbitrary-routed tendons, and simplify it under the assumption of no external forces and torques to be used as a kinematic model, only considering the forces and torques provided by the tendon tensions.

%
%
Motion planning enables robot trajectories that avoid unintentional collision with anatomical structures.
Sampling-based motion planning methods connect randomly sampled robot configurations to generate collision-free paths.
This includes Rapidly-exploring Random Trees (RRT)~\cite{LaValle1998_Report} and Probabilistic Roadmaps (PRM)~\cite{Kavraki1996_TRA}, which construct a collision-free tree and graph, respectively, in the robot's configuration space.
%
There are many variants of RRT and PRM.
RRTConnect grows trees from both the start and goal configurations, frequently attempting to connect them together~\cite{Kuffner2000_ICRA}.
Some variants, such as LazyPRM, employ laziness to improve planning performance, which defers collision checks until after a candidate path is found, updating and replanning if the candidate path is in collision~\cite{Bohlin2000_ICRA}.
PRM$^*$~\cite{Karaman2011_IJRR} and LazyPRM$^*$~\cite{Hauser2015_ICRA} improve roadmap quality by increasing the number of edge connections per sample as the number of samples increases.
Our planner utilizes the connection strategy of PRM$^*$ and laziness on some added edges.

%
%
Motion planning specific to tendon-driven robots has typically been applied to robots with multiple constant curvature segments~\cite{Deng2019_RoboSoft,Zhang2021_JAIT,Meng2021_ICRA,Lai2022_RAL}, with no attention given to planning for tendon-driven robots with arbitrarily-routed tendons.

Fast and interactive motion planning has been explored in many other domains, even with continuum medical robots, such as for steerable needles and concentric-tube robots.
Many of these works employ fast replanning as part of their robot control loop, showing that a fast planner can also be utilized during control of complex continuum robots~\cite{Sun2015_TRO,Patil2014_TRO,Li2018_MBEC,Leibrandt2017_RAM}.
Previous work with concentric-tube robots enabled fast interactive-rate motion planning in anatomical environments utilizing a pre-generated collision-free roadmap, created before-hand using the segmented collision-free environment~\cite{Torres2014_ICRA,Torres2015_ICRA,Leibrandt2017_RAM}.
We build upon these works, presenting a method which plans motions at interactive rates specifically for tendon-driven robots with arbitrarily complex tendon routing and which, in contrast to these methods, does not depend on prior knowledge of the anatomical environment.

\section{Problem Definition}
\label{sec:problemdef}

%
%

We consider a tendon-driven robot of length $\ell$, radius $\rho$, and with $N$ (potentially non-linearly routed) tendons, each with independently controlled tension.
We consider two additional degrees of freedom: rotation of the entire robot about the robot's base, and insertion or retraction of the robot in the environment.
The configuration space is $Q \subseteq \mathds{R}_{\geq 0}^N \times SO(2) \times \mathds{R}_{\geq 0}$, with each configuration representing tendon tensions, rotation about the robot base, and inserted length of the robot into the physical environment.

We define the function $\vec{T}_{\text{tip}} : Q \rightarrow \mathds{R}^3$ as the mapping from a robot configuration to the associated robot tip position, specifically the distal position of the tendon-robot's backbone centerline.
We define two predicates, $\collisionFree(\vec{q})$ and $\selfCollisionFree(\vec{q})$, that evaluate to true if the robot shape at configuration $\vec{q}$ does not collide with the environment or with itself, respectively.
Using these predicates, we define
\begin{align*}
  Q_{\text{valid}}
    &=
      \bigl\{
        \vec{q} \in Q ~|~ \selfCollisionFree(\vec{q})
      \bigr\}
  \\
  Q_{\text{free}}
    &=
      \bigl\{
        \vec{q} \in Q_{\text{valid}} ~|~ \collisionFree(\vec{q})
      \bigr\}
\end{align*}
with $Q_{\text{free}} \subseteq Q_{\text{valid}} \subseteq Q$, where $Q_{\text{valid}}$ consists of configurations free of self-collision and $Q_{\text{free}}$ consists of configurations free of collisions with the environment and itself.
Each of these configuration spaces have respective reachable end-effector workspaces $W_{\text{free}} \subseteq W_{\text{valid}} \subseteq W$, through the $\vec{T}_{\text{tip}}$ transform (e.g., $W = \{ \vec{T}_{\text{tip}}(\vec{q}) ~|~ \vec{q} \in Q \}$).

%
%

Our \textit{motion-planning problem} is to generate a continuous collision-free path $\vec{\sigma}_i : [0, 1] \rightarrow Q_{\text{free}}$ from a given start configuration $\vec{q}_{0} \in Q_{\text{free}}$ to a desired goal robot tip position $\vec{g} \in \mathds{R}^3$.
Note that $\vec{g}$ may not be within the robot tip's collision-free workspace $W_{\text{free}}$.
This motion-planning problem can be formulated as the following constrained optimization
\begin{equation}
  \begin{aligned}
    \argmin_{\vec{\sigma}(s)} \;
      &
        \bigl\|
          \vec{T}_{\text{tip}}
          \big(
            \mspace{-2mu}
            \vec{\sigma}(1)
            \mspace{-2mu}
          \big)
          - \vec{g}
        \bigr\|
    \\
    \text{s.t.} \quad
      &
        \vec{\sigma}(s) \in Q_{\text{free}}, \quad \forall s \in [0, 1]
    \\
    \text{and} \quad
      &
      \vec{\sigma}(0) = \vec{q}_{0}.
  \end{aligned}
\label{eqn:minimize-tip-error}
\end{equation}
This formulation only constrains the path to be collision-free and optimizes the path's reached destination to be close to the goal tip position.
We define
$\vec{q}_{\text{reached}} = \vec{\sigma}(1)$ and
$\vec{g}_{\text{reached}} = \vec{T}_{\text{tip}}(\vec{q}_{\text{reached}}) \in W_{\text{free}}$
as the reached configuration and end-effector destination.

%
%

We define the \textit{supervisory-control problem} as a streaming sequence of motion-planning problems.
Given a streaming sequence of goal robot tip positions, $(\vec{g}_1, \vec{g}_2, \ldots, \vec{g}_i, \ldots)$ with $\vec{g}_i \in \mathds{R}^3$, the supervisory-control problem is to generate a streaming sequence of connected continuous collision-free paths $(\vec{\sigma}_1, \vec{\sigma}_2, \ldots, \vec{\sigma}_i, \ldots)$.
Each generated path $\vec{\sigma}_i$ comes from solving the motion-planning problem in~\eqref{eqn:minimize-tip-error} with $\vec{g}_i$ as the goal.
The first path starts at a given initial robot configuration $\vec{\sigma}_0(0) = \vec{q}_{\text{start}} \in Q_{\text{free}}$.
Each subsequent path starts where the previous one ended, i.e., $\vec{\sigma}_i(0) = \vec{\sigma}_{i-1}(1)$

%
%

We define supervisory control with streaming input goals and streaming output paths to enable interactive use by a human in-the-loop.
Therefore we desire a planning rate capable of human interaction, which we call \textit{interactive-rate}.
In this work, we define the interactive-rate planning time as less than one second (with faster speeds being more useful), as measured between receiving $\vec{g}_i$ and returning the path $\vec{\sigma}_i$.

\section{Method}
\label{sec:method}

To solve this problem, we introduce a motion-planning method divided into three phases, Precompute, Load, and Supervisory-Control as seen in Fig.~\ref{fig:fig1}.
To enable such planning at interactive rates, we introduce a fast kinematic model and collision-checking framework.

\subsection{Efficient Cosserat Rod and Tendon Model}
\label{sec:method:fk}

First, we briefly present the mechanical model for tendon-actuated robots from Rucker \etal~\cite{Rucker2011_TRO} that builds upon the Cosserat rod and string theories.
We then present our method to more efficiently evaluate this model, specifically when there are no external loads on the robot, as in our application.
This extension significantly improves the computational speed of the model.

The model describes the shape of a tendon actuated robot by a framed curve in space through the center,
$\vec{p}(s) \in \mathds{R}^3$
and the orientation
$R(s) \in SO(3)$
as a function of reference arc length
$s \in [0, \ell]$
along the undeformed robot backbone.
We define $\vec{v}(s)$ and $\vec{u}(s)$ as the linear and angular rates of change of $\vec{p}(s)$ and $R(s)$ such that
\begin{equation}
  \dot{R}(s) = R(s) \Sx(\vec{u}(s)),
  \quad
  \dot{\vec{p}}(s) = R(s) \vec{v}(s)
  \label{eq:pR}
\end{equation}
where $\Sx(\cdot)$ maps $\mathds{R}^3$ to $\mathfrak{so}(3)$, producing a skew-symmetric matrix.

The tendons that actuate the Cosserat rod are modeled as Cosserat strings, traveling through predefined routing channels with zero friction, and attached to some location along the backbone, usually at the tip.
The $i^{\text{th}}$ tendon pulls with tension $\tau_i$ at the base.
The tendon routing is described as a two-dimensional parameterized vector
$\vec{r}_i(s) = [x_i(s)\ \ y_i(s)\ \ 0]^\top$
expressed in the backbone reference frame.
As the robot is actuated, the curve describing the path of the  $i^{\text{th}}$ tendon is then
\begin{equation}
  \vec{p}_i(s)
    =
      R(s) \vec{r}_i(s) + \vec{p}(s).
\end{equation}
In \cite{Rucker2011_TRO}, the authors apply the static equilibrium equations for Cosserat rods and strings to determine the forces and moments that the strings apply to the backbone as a function of the applied tensions $\tau_i$.
This ultimately results in the following differential equations defining $\vec{u}$ and $\vec{v}$.
\begin{equation}
\left[
  \begin{array}{c}
    \dot{\vec{v}} \\
    \dot{\vec{u}}
  \end{array}
\right]
  =
    \left[
      \begin{array}{cc}
        K_{se} + A  &  G           \\
        B           &  K_{bt} + H
      \end{array}
    \right]^{-1}
    \left[
      \begin{array}{c}
        \vec{d} \\
        \vec{c}
      \end{array}
    \right]
    \label{eq:uv}
\end{equation}
where $K_{se}$, and $K_{bt}$, are stiffness matrices associated with the constitutive material law and geometry of the backbone, $A$, $B$, $G$, and $H$ are all functions of the tendon paths, the tensions, the distributed external forces $\vec{f}_e(s)$ and torques $\vec{l}_e(s)$, and the point force $\vec{F}_e$ and torque $\vec{L}_e$ at the tip of the robot.
The variables $\vec{c}$ and $\vec{d}$ are functions of those quantities as well, but also functions of $\vec{u}$ and $\vec{v}$ respectively.
These expressions are provided in detail in~\cite{Rucker2011_TRO}.

In the general case with external loads on the robot, the above model is a two-point boundary-value problem (BVP) in which $\vec{p}(0)$ and $R(0)$ are known, and static equilibrium conditions must be satisfied at the robot tip for the internal backbone force $\vec{n}(\ell)$ and internal moment $\vec{m}(\ell)$.
In \cite{Rucker2011_TRO}, the problem is solved by any nonlinear 
multivariable root finding algorithm to iteratively update $\vec{v}(0)$ and $\vec{u}(0)$ such that
\begin{equation}
  \begin{aligned}
    \vec{n}(\ell) - \vec{F}_t - \vec{F}_e &= \vec{0} \\
    \vec{m}(\ell) - \vec{L}_t - \vec{L}_e &= \vec{0}
  \end{aligned}
  \label{eqn:uv-root-finding}
\end{equation}
where $\vec{F}_t$ and $\vec{L}_t$ are the force and torque at the tip due to the tendon tensions, and $\vec{n}(\ell)$ and $\vec{m}(\ell)$ are computed using numerical integration of~\eqref{eq:pR} and~\eqref{eq:uv}.
Typically, an iterative optimization algorithm, like Levenberg-Marquardt (LM)~\cite{Nocedal2006_Book_ch10}, is employed to update initial guesses of $\vec{v}(0)$ and $\vec{u}(0)$ and through an integration along the robot backbone, evaluate the distal force and torque boundary conditions.

In this work, to facilitate rapid motion planning, we develop an alternative, more efficient way to solve the tendon-actuated robot model in the special case of no external loads (i.e., for $\vec{f}_e(s) = \vec{0}$, $\vec{l}_e(s) = \vec{0}$, $\vec{F}_e = \vec{0}$, and $\vec{L}_e = \vec{0}$).
By considering the entire collection of robot backbone and tendons from $s=0$ to $s=\ell$ as a single body, we can write a force and moment balance at the robot's base, resulting in
\begin{equation}
  \begin{aligned}
    \vec{n}(0)
      &=
        -
        \sum_{i=1}^N
        \tau_i
        \frac{
          \dot{\vec{p}}^b_i(0)
        }{
          \| \dot{\vec{p}}^b_i(0) \|
        }
    \\
    \vec{m}(0)
      &=
        -
        \sum_{i=1}^N
        \tau_i
        \frac{
          \vec{r}_i(0)
          \times
          \dot{\vec{p}}^b_i(0)
        }{
          \| \dot{\vec{p}}^b_i(0) \|
        }
  \end{aligned}
\label{eq:eq}
\end{equation}
when expressed in body-frame coordinates, where
\begin{equation}
\label{eq:pidot}
  \dot{\vec{p}}^b_i(0)
    \equiv
      R(0) \dot{\vec{p}}_i(0)
    =
      \vec{u}(0)
      \times
      \vec{r}_i(0)
      +
      \dot{\vec{r}}_i(0)
      +
      \vec{v}(0).
\end{equation}
The internal force and moment are also related to $\vec{u}(0)$ and $\vec{v}(0)$ through the backbone material constitutive law as
\begin{equation}
\begin{aligned}
\label{eq:const}
  \vec{v}(0)
    &=
      K_{se}^{-1} \vec{n}(0)
      +
      [0\ 0\ 1]^\top
  \\
  \vec{u}(0)
    &=
      K_{bt}^{-1} \vec{m}(0).
\end{aligned}
\end{equation}
The above nonlinear algebraic system can be numerically solved to obtain the correct $\vec{u}(0)$ and $\vec{v}(0)$ before the differential equations are ever integrated.
Any root-finding algorithm could be used, but we use a simple fixed-point iteration approach that converges quickly (i.e. choose a $\vec{u}(0)$ and $\vec{v}(0)$, evaluate $\dot{\vec{p}}^b_i(0)$ using~\eqref{eq:pidot}, evaluate the force and moment in~\eqref{eq:eq}, and then update $\vec{u}(0)$ and $\vec{v}(0)$ using~\eqref{eq:const}).
After convergence, this results in a full set of initial conditions that enables us to subsequently numerically integrate the model equations~\eqref{eq:pR} and~\eqref{eq:uv} to obtain the robot shape and end-effector pose.
See Appendix~\ref{apdx:fk} for implementation details.

Compared to the prior general approach of solving the two-point BVP, this approach is exceptionally advantageous for efficient motion planning because it converts the problem into an initial-value problem (IVP) which can be evaluated in far less time.
The entire model evaluation involves solving the small nonlinear problem in~\eqref{eq:const} and then a single integration along the robot length of the model differential equations in~\eqref{eq:uv} with each integration step requiring constructing and solving a small linear equation of six variables.
In our experiments, we compare our fast IVP formulation against a standard implementation of the shooting method to solve the BVP, which wraps the forward integration with an outer optimization loop to satisfy the boundary conditions.

Along with the tendon tensions, we enable two more configuration controls, rotation and extension/retraction of the robot about the point of insertion into the environment.
Retraction for a physical robot may be handled by the base of the robot pulling the robot in and out of the point of insertion with a rigid but collapsing sheath constraining the robot proximal to the insertion point.
The retraction degree of freedom is modeled by performing the shape computation starting from the point of the robot that is at the point of insertion.
Having a high value of retraction quickens the shape computation because we integrate over a shorter length.
The rotation about the insertion point is modeled by a rotation about the $z$-axis (as in Fig.~\ref{fig:tendon-robot}) after performing the shape computation.
These additional degrees of freedom enable more expressive movements without significant computational overhead.

We leverage our fast FK solver to compute the shape of the robot for detecting collisions during motion planning and to enable IK in reaching desired robot tip positions.

\subsection{Collision Checking}

Collision checks involve detecting robot self-collisions and collisions between the robot and the environment.
The tendon-robot's geometry is fully described by the robot's radius $\rho$, swept along the robot's backbone, $\vec{p}(s)$.
Our FK solver produces a finely discretized piecewise-linear approximation of the backbone's continuous shape using numerical integration of the kinematic model.
Utilizing this piecewise-linear backbone, we fully specify the shape of the robot using a sequence of capsules (cylinders with spherical caps), resulting in a constant $\rho$ radius around the discretized backbone shape.

\subsubsection{Self-Collision Detection}

Self-collision occurs when the robot bends around itself, causing the surface of one robot segment to intersect with the surface of another robot segment.
For detecting self-collision, we explicitly represent the sequence of capsules along the computed backbone shape.
To avoid false positives for adjacent capsules, we collision-check each capsule with all other capsules that are further than $3\rho$ away along the backbone, in order to avoid false positives for adjacent capsules.
Capsules are checked analytically by determining the closest points between the two line segments, then checking if their distance is less than the robot diameter.

\subsubsection{Environmental Collision Detection via Voxels}

%
%

We note that segmenting pre-operative medical images, such as Computed Tomography (CT) and Magnetic Resonance Imaging (MRI) scans, naturally produces voxel-based anatomical representations.
We here present a method that leverages this representation to enable computationally efficient collision detection between the robot and the patient's anatomy.
We do so by voxelizing the robot's geometry and performing efficient voxel-to-voxel collision detection between the robot and the anatomical segmentation.

%
%

We first recognize that the robot shape represents a dilation of radius $\rho$ from the robot backbone's centerline.
The voxelization of the full robot shape has a high computational cost as it requires identifying all voxels within radius $\rho$ away from the robot backbone centerline.
To avoid this high computational cost repeatedly in planning, we instead collision-check the robot backbone's centerline directly against the obstacle environment after dilating the environment by a ball of radius $\rho$, which is equivalent.
We voxelize the backbone centerline shape via the efficient algorithm presented in~\cite{Amanatides1987_EG}.
The dilation of the environment voxel representation is only performed once per environment.
This enables efficient voxelization of the robot's backbone centerline shape for detecting collisions during planning.

%
%

The two key insights behind our fast voxel-to-voxel collision detection method are (i) the set of occupied voxels is sparse when considering the full imaging volume, particularly for the robot's backbone, and (ii) a voxel segmentation is effectively a boolean occupancy map, enabling the use of extremely fast bitwise computational operators for collision detection.

We store the voxel sets in $\mathds{R}^3$ as sparse voxel octrees, where the space is structured as a tree with a maximum branching factor of eight, one for each of the sub-octants of the $\mathds{R}^3$ environment~\cite{Meagher1982_CGIP}.
If a sub-octant has no occupied voxels, it can be safely discarded, meaning a missing child represents unoccupied space, resulting in an efficient sparse representation.
At the leaves of our octree we represent a block of \num{4 x 4 x 4} voxels as a \SI{64}{\bit} unsigned integer with each bit representing a single voxel.
Using this representation, we leverage efficient bitwise operations to perform collision detection (bitwise AND), and to perform unions (bitwise OR).
Due to the sparsity, operations between two voxel sets, like collision detection, can skip large subtrees if it is empty in either of the two voxel sets.
Collision checking recurses down both voxel octrees, only checking the leaf blocks that are occupied by both octrees.
Efficient collision detection between two leaf blocks, each represented by a \SI{64}{\bit} unsigned integer, amounts to checking if the AND operation between their values is non-zero.

%
%

A key component in roadmap-based motion planning is the collision detection of, not only single robot configurations, but also the edges representing motions the robot takes between connected configurations.
To address this, we use the voxelization of the volume swept by the robot's geometry as it moves in $Q_{\text{valid}}$ from one configuration to another.
Typically, detecting collisions for motions involve individual collision checks at equally-spaced discretizations along the motion's path in configuration space.
The main difficulty with this approach lies in determining an appropriate minimum configuration distance for the motion discretization.
If we discretize too closely, we pay with excessive computations, but if we discretize too broadly, we may have gaps in the discretization, potentially resulting in undetected collision.
High nonlinearities of the tendon-driven robot further aggravate the situation---some small configuration changes result in small changes to the robot shape, but others result in very large robot shape deformations.
We further leverage our voxel representation to efficiently and dynamically compute the swept volume such that there are no gaps up to the resolution of the voxels themselves.
This dynamic algorithm recursively subdivides the motion until each adjacent robot backbone shape is less than or equal to one voxel distance away for each backbone point.

For our dynamic recursive subdivision, we define the distance between two configurations in voxel space as
\begin{equation}
  \max_{s}
  \left\|
    \voxelIdx(\vec{p}_a(s))
    -
    \voxelIdx(\vec{p}_b(s))
  \right\|_\infty
  \label{eqn:voxel-distance}
\end{equation}
where $\vec{p}_a(s)$ and $\vec{p}_b(s)$ are the backbone centerline shapes parameterized by path length of the two endpoint configurations of a motion interval, and where $\voxelIdx(\cdot) : \mathds{R}^3 \rightarrow \mathds{N}^3$ converts a point in Euclidean space into 3D voxel indices.
Our dynamic algorithm subdivides an interval if its configuration-space distance is larger than a user-specified threshold and if the voxel distance from~\eqref{eqn:voxel-distance} is larger than one.
We have another edge discretizer variant that voxelizes from the first configuration until collision, in which case, we add an additional requirement for subdivision that  a voxelized $\vec{p}_a(s)$ does not collide with the environment.
We provide more detail on these algorithms in Appendix~\ref{apdx:edge-voxelization}.

Depending on the discretization used for the constant stepping of the traditional approach, our dynamic algorithm will either be more computationally efficient by avoiding unnecessary subdivisions, or will be more accurate and correct, resulting in a voxelized swept volume without gaps.

\subsection{Interactive Supervisory Control}

The fast FK and collision-detection methods above enable our motion-planning method to achieve interactive-rate supervisory control.

Our interactive supervisory-control method has three phases, Precompute, Load, and Supervisory-Control (see Fig.~\ref{fig:fig1}).

\subsubsection{Precompute Phase}

%
%

In the Precompute phase, we generate a large roadmap which encodes movement of the tendon-driven robot in free space.
The roadmap is an undirected graph consisting of a set of vertices $V$ and a set of edges $E$.
Each vertex represents a single tendon-robot configuration from $Q_{\text{valid}}$, meaning it is free of self-collision and lies within configuration limits.
Each edge connects two vertex configurations and represents motion between the vertices via linear interpolation in configuration space.

We first sample $N_r$ configuration vertices from $Q_{\text{valid}}$.
These samples are then connected by edges to their $k_n$ nearest neighbors in configuration space.
We choose $k_n$ according to the PRM$^*$ algorithm~\cite{Karaman2011_IJRR}.
We then voxelize the configurations at each vertex and the swept volumes at each edge, removing edges that exit $Q_{\text{valid}}$ (i.e., motions that self-collide or violate configuration limits).
This ordering of sampling, then connecting, then voxelizing enables trivial parallelizability for precomputing the roadmap.

In order to ensure a sufficiently expansive roadmap after pruning away in-collision vertices and edges, which happens in the Load phase, we require a dense roadmap that covers the robot's workspace as much as possible.

%
%

\begin{figure}
  \centering
  \includegraphics[width=\columnwidth]{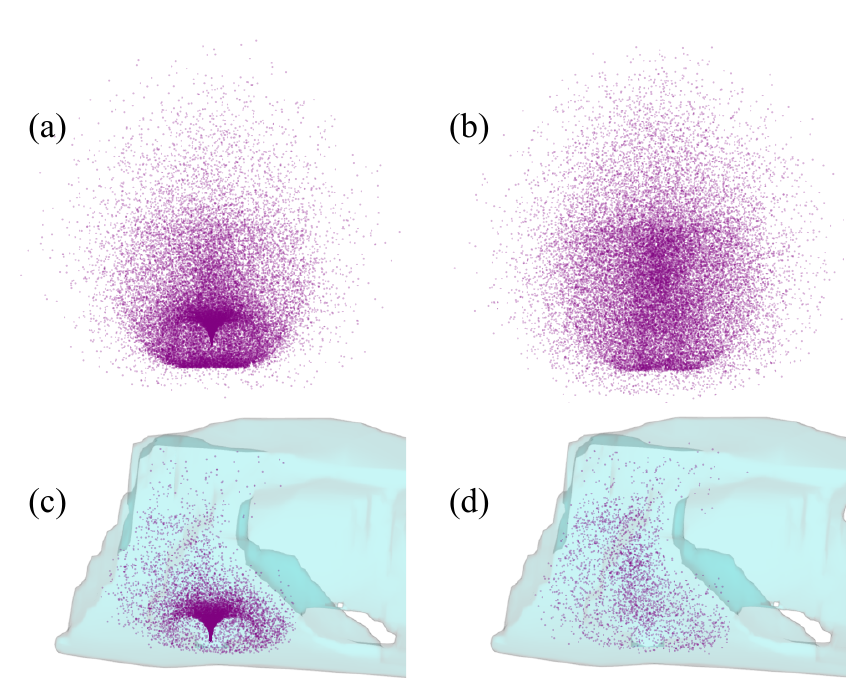}
  \caption{
    A comparison of tip positions of \num{30 000} samples from sampling strategies with a tendon robot capable of retraction.
    The only difference in sampling is for the retraction degree of freedom.
    (a) shows na\"ive uniform sampling of retraction as $s = \ell u$ where $u \sim U(0, 1)$.
    (b) shows spherical sampling of retraction as $s = \ell \sqrt[3]{u}$ where $u \sim U(0, 1)$.
    (c) and (d) are after pruning from our method's Load phase of (a) and (b), respectively.
  }
  \label{fig:config-sampling}
\end{figure}

Usually, random configurations are chosen uniformly from the configuration space.
However, Fig.~\ref{fig:config-sampling} shows uniform random sampling in our configuration space causes many configurations to be sampled near the insertion point.
This high concentration of samples decreases the quality of our roadmap and leaves less expressiveness toward configurations with high robot extension values.
We overcome this limitation with a non-uniform sampling strategy.
Since the reachable workspace of the robot is restricted within a ball of radius $\ell$ centered at the origin, we sample the retraction degree of freedom equivalently to the radius component of a uniformly sampled ball in $\mathds{R}^3$, which is simply $\ell \sqrt[3]{u}$ where $u \sim U(0,1)$~\cite{Voelker2017_Report} (a derivation is provided in Appendix~\ref{apdx:spherical-sampling}).
Fig.~\ref{fig:config-sampling} demonstrates this sampling strategy.
Using this sampling for the extension degree of freedom, we achieve a much more distributed roadmap in tip position space.

%
%

In addition to caching the voxelized vertices and edges, the robot tip position at each vertex state is also cached into the roadmap for later use by the IK solver to find nearest neighbors in tip-position space.
This precomputation needs to be computed once per robot and can then be reused in many procedures and with various anatomical environments.

\subsubsection{Load Phase}

%
%

To enable the robot to move in a specific anatomical environment effectively, we take as input a voxel-based segmentation of the patient's anatomy and the free-space in which the robot will be operating.
This can come from, e.g., a pre-operative volumetric image of a cavity in the patient's body and the surrounding anatomy.

We first process and modify the given segmented anatomical space.
We do so by dilating the boundary of the free space, i.e., the surrounding anatomy, expanding each voxel to fill neighboring voxels up to the robot radius in distance.
This effectively shrinks the available free space inside the anatomy so that we can safely check collisions only against the voxelized robot backbone centerline.

We next load the precomputed roadmap and perform voxel-based collision detection between the roadmap (vertices and edges) and the anatomical segmentation.
We discard any vertex or edge in the roadmap found to be in collision with the environment.

After removing vertices and edges in collision, we ensure that the graph contains only one connected component by further removing all vertices that are not part of the main connected component, since those disconnected vertices are unreachable from the main connected component.

\subsubsection{Supervisory-Control Phase}

After loading the roadmap and pruning in-collision vertices and edges, our method is ready to plan interactively to a stream of desired tip positions $\{\vec{g}_1, \vec{g}_2, \ldots, \vec{g}_i, \ldots\}$.
The pruned roadmap resulting from the Load phase provides the initial internal roadmap representation for the planning in this phase.
The method is based on LazyPRM, except that most vertices and edges are precomputed and explicitly collision-checked instead of lazily evaluated, e.g., during the Load phase.
However some edges to newly added goal vertices during the Supervisory-Control phase are lazily evaluated.

\begin{algorithm}[t]
  \caption{Supervisory Control}
  \label{alg:supervisory-control}

  \DontPrintSemicolon
  \SetArgSty{} 
  \SetDataSty{} 
  \SetKwInOut{Input}{input}
  \SetKwInOut{Output}{output}
  \SetKwComment{tcp}{\textbf{//} }{}
  \newcommand{\myAlgCommentSty}[1]{\color{blueishgreen}#1}
  \SetCommentSty{myAlgCommentSty}

  \SetKwRepeat{Do}{do}{while}
  \SetKwData{threshold}{threshold}

  \KwIn{\\
    \quad $\vec{q}_{\text{start}}$:  from $Q_{\text{free}}$, initial robot configuration\\
    \quad $(V, E)$:                  pruned roadmap vertices and edges;\\
    \quad $k_{\text{IK}}$:           number of tip-space neighbors for \RoadmapIK\\
    \quad \threshold:                tip-error threshold
  }

  $\vec{q}_{\text{current}} \leftarrow \vec{q}_{\text{start}}$ \;
  \While{procedure is not done}{
    $\vec{g} \leftarrow$ get goal tip-position from the user \;
    \tcp{Note: \RoadmapIK (Alg.~\ref{alg:roadmap-ik}) adds to $(V, E)$}
    $\vec{q}_{\text{reached}}
      \leftarrow
        \RoadmapIK((V, E), k_{\text{IK}}, \vec{g}, \threshold)\!\!\!$
    \;
    \Do{any lazy edge in $\vec{\sigma}$ collides with anatomy}{
      $\vec{\sigma}
        \leftarrow$
          A$\!^*$ on $(V, E)$
          from $\vec{q}_{\text{current}}$
          to $\vec{q}_{\text{reached}}$
      \;
    }
    run plan $\vec{\sigma}$ on robot \;
    $\vec{q}_{\text{current}} \leftarrow$ last configuration in $\vec{\sigma}$ \;
  }
\end{algorithm}

We detail the Supervisory-Control phase in Alg.~\ref{alg:supervisory-control}.
At a high level, the method begins as the clinician specifies a desired tip position, e.g., via the use of a haptic teleoperation device.
Our method first determines an IK solution that places the robot's tip at or near the desired location.
It then inserts the IK solution into the existing roadmap and generates a plan on the roadmap, corresponding to a sequence of collision-free motions from the robot's current configuration $\vec{q}_{\text{current}}$ to the IK solution.
This plan is then executed on the robot.
The method then waits for the clinician to specify the next desired tip position and we repeat the process until the clinician has completed the procedure.

%
%

\begin{algorithm}[t]
  \caption{\RoadmapIK}
  \label{alg:roadmap-ik}

  \DontPrintSemicolon
  \SetArgSty{} 
  \SetDataSty{} 
  \SetKwInOut{Input}{input}
  \SetKwInOut{Output}{output}
  \SetKwComment{tcp}{\textbf{//} }{}
  \newcommand{\myAlgCommentSty}[1]{\color{blueishgreen}#1}
  \SetCommentSty{myAlgCommentSty}
  \SetFuncSty{normaltext}

  \SetKwFunction{IKsolve}{IK}
  \SetKwData{threshold}{threshold}
  \SetKwData{None}{None}
  \SetKw{KwBreak}{break}
  \SetKw{KwIs}{is}

  \KwIn{\\
    \quad $(V, E)$:        roadmap vertices and edges; $V \subseteq Q_{\text{free}}$ \\
    \quad $k_{\text{IK}}$: number of tip-space neighbors to use with IK \\
    \quad $\vec{g}$:       from $\mathds{R}^3$, goal tip position \\
    \quad \threshold:      tip error threshold
  }
  \KwOut{$\vec{q}_{\text{reached}}$: from $Q_{\text{free}}$, IK reachable from the roadmap}

  $Q_{\text{nn}}
    \leftarrow
      \vec{g}$'s nearest $k_{\text{IK}}$ neighbors from $V$ in tip-space \;
  $E_{\text{unchecked}} \leftarrow \emptyset$ 
    \tcp*[r]{unchecked IK edges}
  $E_{\text{reached}} \leftarrow \emptyset$ 
    \tcp*[r]{evaluated IK edges}
  $e_{\text{found}} \leftarrow \None$
    \tcp*[r]{found edge $<$ threshold (or closest)}
  \ForEach{$\vec{q}_{\text{nn}} \in Q_{\text{nn}}$}{
    $\vec{q}_{\text{IK}}
      \leftarrow
        \IKsolve(\vec{q}_{\text{nn}}, \vec{g}, \threshold)$
        \tcp*[r]{using LM}
    \tcp{check for early success}
    \If{$\|\vec{T}_{\text{tip}}(\vec{q}_{\text{IK}}) - \vec{g}\| < \threshold$}{
      $\vec{q}_{\text{reached}}
        \leftarrow
          \VoxelizeFreeEdge(\vec{q}_{\text{nn}}, \vec{q}_{\text{IK}})$ \;
          \label{alg:roadmap-ik:partial-edge-1}
      \If{$\|\vec{T}_{\text{tip}}(\vec{q}_{\text{reached}}) - \vec{g}\| < \threshold$}{
        $e_{\text{found}}
          \leftarrow
            (\vec{q}_{\text{nn}}, \vec{q}_{\text{reached}})$
          \;
          \KwBreak \tcp*[r]{exit the \textbf{foreach} loop}
      }
      add $(\vec{q}_{\text{nn}}, \vec{q}_{\text{reached}})$
        to $E_{\text{reached}}$ \;
    } \Else {
      add $(\vec{q}_{\text{nn}}, \vec{q}_{\text{IK}})$
        to $E_{\text{unchecked}}$ \;
    }
  }
  \If(\tcp*[f]{find closest instead}){$e_{\text{found}}$ \KwIs \None}{
    \ForEach(\tcp*[f]{eval the rest}){
        $(\vec{q}_{\text{nn}}, \vec{q}_{\text{IK}}) \in E_{\text{unchecked}}$
    }{
      $\vec{q}_{\text{reached}}
        \leftarrow
          \VoxelizeFreeEdge(\vec{q}_{\text{nn}}, \vec{q}_{\text{IK}})$ \;
          \label{alg:roadmap-ik:partial-edge-2}
      add $(\vec{q}_{\text{nn}}, \vec{q}_{\text{reached}})$ to $E_{\text{reached}}$ \;
    }
    $e_{\text{found}}
      \leftarrow
        \displaystyle
        \argmin_{(\cdot, \vec{q}_{\text{reached}}) \in E_{\text{reached}}}
        \|\vec{T}_{\text{tip}}(\vec{q}_{\text{reached}}) - \vec{g}\|$ \;
  }
  $(\vec{q}_{\text{nn}}, \vec{q}_{\text{reached}})
    \leftarrow
      e_{\text{found}}
  $ \;
  add $\vec{q}_{\text{reached}}$ to $V$ and checked edge $e_{\text{found}}$ to $E$ \;
    \label{alg:roadmap-ik:add-known-edge}
  add PRM$^*$ lazy edges to $E$ (from $\vec{q}_{\text{reached}}$
    \newline
    to $\vec{q}_{\text{reached}}$'s nearest neighbor configs in $V$) \;
    \label{alg:roadmap-ik:add-prm-edges}
  \Return $\vec{q}_{\text{reached}}$ \;
\end{algorithm}

We frame the IK problem as an optimization similar to~\eqref{eqn:minimize-tip-error}, finding a collision-free robot configuration that minimizes the distance between the predicted and desired robot tip positions.
\begin{equation}
  \argmin_{\vec{q}_{\text{reached}} \in Q_{\text{free}}} \;
    \bigl\|
      \vec{T}_{\text{tip}}
      \big(
        \mspace{-2mu}
        \vec{q}_{\text{reached}}
        \mspace{-2mu}
      \big)
      - \vec{g}
    \bigr\|
\label{eqn:ik-problem}
\end{equation}
We utilize the Levenberg-Marquardt (LM) optimization algorithm to iteratively update an initial configuration guess until convergence to the desired tip position, or further progress is not possible~\cite{Nocedal2006_Book_ch10}.
Use of LM is commonly performed for IK of robots with redundant manipulators~\cite{Shunsuke2012_JRM}, but as a gradient-based optimizer, the effectiveness of LM depends heavily on the given initial guess.
Our \RoadmapIK algorithm (Alg.~\ref{alg:roadmap-ik}) leverages the motion-planner's roadmap to extract configurations that have robot tip positions nearest to the desired goal $\vec{g}$ to be used as initial guesses for LM.
However, with the nonlinear relationship between configuration space and robot tip positions, robot configurations with nearby tip positions may be very distant in configuration space, especially when considering obstacles.
Therefore, we allow \RoadmapIK to execute multiple LM instances, each starting from one of $k_{\text{IK}}$ of such nearest configurations, returning either the first to successfully reach $\vec{g}$ or the closest IK across all LM instances.

There are two problems with directly using LM to obtain a goal configuration for use in generating a collision-free path through the roadmap.
First, the LM algorithm can employ linear constraints such as configuration limits, but cannot handle nonlinear constraints such as collision avoidance.
If $\vec{q}_{\text{IK}}$, the solution returned from LM, does not belong to $Q_{\text{free}}$, then it must somehow be projected onto $Q_{\text{free}}$ to satisfy constraints in~\eqref{eqn:ik-problem} and~\eqref{eqn:minimize-tip-error}.
Second, even if $\vec{q}_{\text{IK}}$ is in $Q_{\text{free}}$, there may not be a collision-free path through the roadmap to $\vec{q}_{\text{IK}}$, either because $\vec{q}_{\text{current}}$ and $\vec{q}_{\text{IK}}$ belong to disconnected regions of $Q_{\text{free}}$ or because our roadmap is not sufficiently refined to provide a collision-free linear edge to $\vec{q}_{\text{IK}}$.

For the second problem, we recognize that our roadmap consists of a single fully-connected component, meaning there exists a collision-free path between any two vertices.
In order to maintain this desirable property, we add the additional constraint on~\eqref{eqn:ik-problem} that $\vec{q}_{\text{reached}}$ can connect to the roadmap with a single collision-free edge.
\begin{equation}
  \begin{aligned}
    \argmin_{\vec{q}_{\text{reached}} \in Q_{\text{free}}} \;
      &
      \bigl\|
        \vec{T}_{\text{tip}}
        \big(
          \mspace{-2mu}
          \vec{q}_{\text{reached}}
          \mspace{-2mu}
        \big)
        - \vec{g}
      \bigr\|
    \\
    \text{s.t.} \quad
      &
      \exists \vec{q} \in V,
    \\
    \text{with} \quad
      &
      (1-s) \vec{q} + s \vec{q}_{\text{reached}} \in Q_{\text{free}},
      \quad
      \forall s \in [0, 1]
  \end{aligned}
\label{eqn:ik-problem-mod}
\end{equation}
With this added constraint, we ensure our roadmap consists of a single fully-connected component throughout its lifetime, which implicitly guarantees the existence of a path from $\vec{q}_{\text{current}}$ to $\vec{q}_{\text{reached}}$.

For the first problem, we can choose a collision-free configuration that lies between the LM solution, $\vec{q}_{\text{IK}}$, and its initial guess $\vec{q}_{\text{nn}}$, one of the nearest neighbor roadmap vertices.
Note that $\vec{q}_{\text{nn}} \in Q_{\text{free}}$ so there exists at least one collision-free configuration between them.
If we solve~\eqref{eqn:ik-problem} with this approach, we would choose the closest collision-free configuration to $\vec{q}_{\text{IK}}$.
However, with the added constraint in~\eqref{eqn:ik-problem-mod}, we choose the closest collision-free configuration to $\vec{q}_{\text{IK}}$ such that it has a collision-free connection to $\vec{q}_{\text{nn}}$.
We find this configuration in the $\VoxelizeFreeEdge()$ function (Alg.~\ref{alg:roadmap-ik} lines~\ref{alg:roadmap-ik:partial-edge-1} and~\ref{alg:roadmap-ik:partial-edge-2}) using our dynamic edge voxelization method which calculates the swept-volume voxel set until a collision and returns the last configuration in $Q_{\text{free}}$ before the collision.
We provide more details of the $\VoxelizeFreeEdge$ algorithm in Appendix~\ref{apdx:edge-voxelization} (Alg.~\ref{alg:voxelize-free-edge}).

Once we determine the closest (or first reaching) IK solution $\vec{q}_{\text{reached}}$ from the $k_{\text{IK}}$ nearest neighbors, we connect $\vec{q}_{\text{reached}}$ to the roadmap with the now known collision-free edge to $\vec{q}_{\text{nn}}$ (Alg.~\ref{alg:roadmap-ik} line~\ref{alg:roadmap-ik:add-known-edge}), which is one of the closest roadmap vertices in terms of tip position, but may be far in configuration space.
In addition to adding this fully evaluated new edge, we add many lazily evaluated edges to $k_n$ nearest neighbors in configuration space (Alg.~\ref{alg:roadmap-ik} line~\ref{alg:roadmap-ik:add-prm-edges}), ensuring we maintain a high quality roadmap.

Our Supervisory Control algorithm (Alg.~\ref{alg:supervisory-control}) then performs A$\!^*$ from the robot's current configuration to the \RoadmapIK solution $\vec{q}_{\text{reached}}$, evaluating any unevaluated edges in the LazyPRM fashion.
The found collision-free path $\vec{\sigma}$ is then executed on the robot, bringing the tip to $\vec{g}_{\text{reached}}$.
We repeat the process as the clinician specifies a new goal tip position.

\section{Evaluation}
\label{sec:evaluation}

We evaluate each of our contributions on a simulated tendon robot within a segmented anatomical environment.
We first describe aspects of our experimental setup.
We then evaluate each of our contributions in depth.
All computation and experimentation were conducted on a computer with \SI{64}{\gibi\byte} of \SI{3200}{\Hz} DDR4 DRAM and an AMD Ryzen 7 3700X 8-core processor.

\subsection{Experimental Setup}

%
%
\subsubsection{Simulated Tendon Robot}

For all experimental evaluations, we consider a simulated tendon robot with length $\ell = \SI{120}{\mm}$, having a radius $\rho = \SI{3}{\mm}$, and with three tendons.
One tendon routes straight along one of the sides and the other two are oppositely-wrapped in a helical shape.
All three tendons maintain a constant distance of \SI{2.5}{\mm} from the center of the backbone.
The helical tendons wrap at a pitch of \SI{0.05}{\radian\per\mm}, resulting in approximately \SI{95}{\percent} of one full revolution around the robot at the full length of \SI{120}{\mm}.
The two oppositely wrapped helical tendons are separated at the base by \SI{180}{\degree}, with the straight tendon entering halfway between them behind where the helical tendons first intersect.
We limit each tendon's tension during actuation between zero and \SI{3.5}{\newton}.

As our physical robot design leverages linear actuators to pull on the tendons, controlling on tendon lengths rather than tensions, we incorporate length limits in addition to tension limits into our configuration limits.
In computing the tendon-robot shape, we also calculate the routed path length of each tendon.
We define tendon length controls as the difference between the tendon path lengths of the current robot shape and the zero-tension robot shape.
We use length limits of \SIrange{-27}{46}{\mm} for the helical tendons and \SIrange{-29}{48}{\mm} for the one straight tendon.
A positive length change pulls out and tightens the tendon, associated with an increase in tension, and a negative length change provides slack to the tendon.
This additional configuration constraint enables a motion plan generated with tensions to be equivalently controlled with tendon lengths.

\subsubsection{Anatomical Environment}

\begin{figure}
  \centering
  \includegraphics[width=\columnwidth]{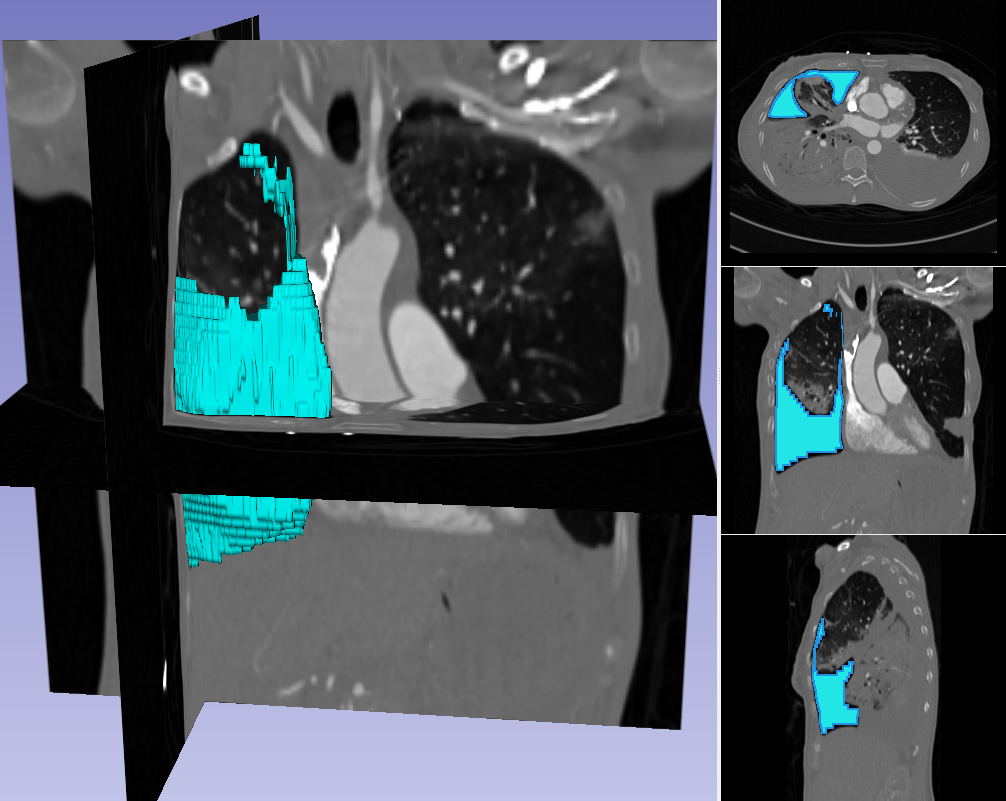}
  \caption{
    Three slices from the original Computed Tomography (CT) scan of the collapsed lung.
    The anatomical environment was segmented from a pleural effusion CT scan using 3D Slicer~\cite{Pinter2019_CMPB}.
  }
  \label{fig:ct-slices}
\end{figure}

In many of our experiments, we simulate the robot moving through an anatomical environment.
This environment corresponds to the pleural space in a patient---the free space between the patient's chest wall and a collapsed lung.
We segmented the pleural space and its boundaries from a CT image of a patient with a collapsed lung, with resolution of \SI{0.59}{\mm} in both $x$ and $y$, and \SI{5}{\mm} in $z$, of size \num[input-product=x]{512 x 512 x 63}, as seen in Fig.~\ref{fig:ct-slices}.
In our experiments, the robot moves through the free space with the boundaries utilized as the surface of the anatomy, with which the robot must avoid colliding.
We utilize both this voxel segmentation in our experiments and a mesh created from the segmentation with \num{3 407} triangle faces for comparison purposes.

As mentioned previously, because of the robot's constant radius, we dilate the voxel environment by the tendon robot radius to simplify collision checking, checking the dilated environment against the voxelized robot backbone centerline.
To avoid over-aggressively dilating the environment, we first subdivide the $z$-dimension by a factor of \num{8}, making each voxel almost a cube, changing the environment to a \num{512 x 512 x 512} voxel grid, each having dimensions \SI{0.59 x 0.59 x 0.625}{\mm}.
We simulate entry of the tendon-driven robot into the pleural space between two ribs identified from the CT scan and consistent with procedures in interventional pulmonology.
See Fig.~\ref{fig:ct-slices} for the CT scan and segmentation.

We restrict all goal tip positions used in evaluation to the region within the dilated environment to ensure the full robot tip would be within the environment if the robot reaches it.
However, all compared methods will still work if given desired goal tip positions within obstacles.

\subsection{Experimental Results}

We first evaluate the speed and accuracy of our fast FK method and compare to existing methods.
We next evaluate our voxel-based collision detection strategy and compare against standard collision detection approaches.
Next, we evaluate our full motion-planning-based supervisory control method against a number of existing motion planners within the anatomical environment on a sequence of desired tip positions.
Finally, to evaluate the influence and contribution of each aspect of our full method, we perform an ablation study.

\subsubsection{Forward Kinematics Study}

We compare our FK model against two versions of a shooting method implementation of the FK model from Rucker \etal~\cite{Rucker2011_TRO} on \num{10 000} randomly generated configurations.
We separately evaluate with and without the retraction degree of freedom enabled.

%
%

As described above, our FK model implementation uses the fixed-point iteration approach in~\eqref{eq:eq} to solve for the initial condition for $\vec{u}(0)$ and $\vec{v}(0)$, converting the BVP into an IVP, and requiring only a single forward integration.

The compared shooting method implementation uses our same forward integration code and solves~\eqref{eqn:uv-root-finding} by minimizing the $\ell_2$ norm of the residual tip forces and torques, with $\vec{F}_e = \vec{0}$ and $\vec{L}_e = \vec{0}$, using LM.
For both our method's fixed point iteration and the shooting method's optimization, We initialize $\vec{u}(0) = [0, 0, 0]^{\top}$ and $\vec{v}(0) = [0, 0, 1]^{\top}$, which are the values of $\vec{u}(0)$ and $\vec{v}(0)$ with zero tension (since our backbone is straight at rest).

%
%

The performance cost of the shooting method implementation is dominated by the number of forward-integration passes.
The LM algorithm utilizes the Jacobian of the FK model with respect to $\vec{u}(0)$ and $\vec{v}(0)$.
We compute this Jacobian numerically using finite-differences which requires repeated forward integrations.
We evaluate this shooting method implementation using forward-differences and central-differences for computing the numerical Jacobian.
Forward-differences requires only seven forward-integration passes, whereas
central-differences requires twelve forward-integration passes.
However, the higher-fidelity Jacobians produced by central-differences may enable better convergence properties, resulting in lower errors and fewer iterations.
See Appendix~\ref{apdx:fk} for implementation details.

\begin{figure*}
  \centering
  \includegraphics[width=\textwidth]{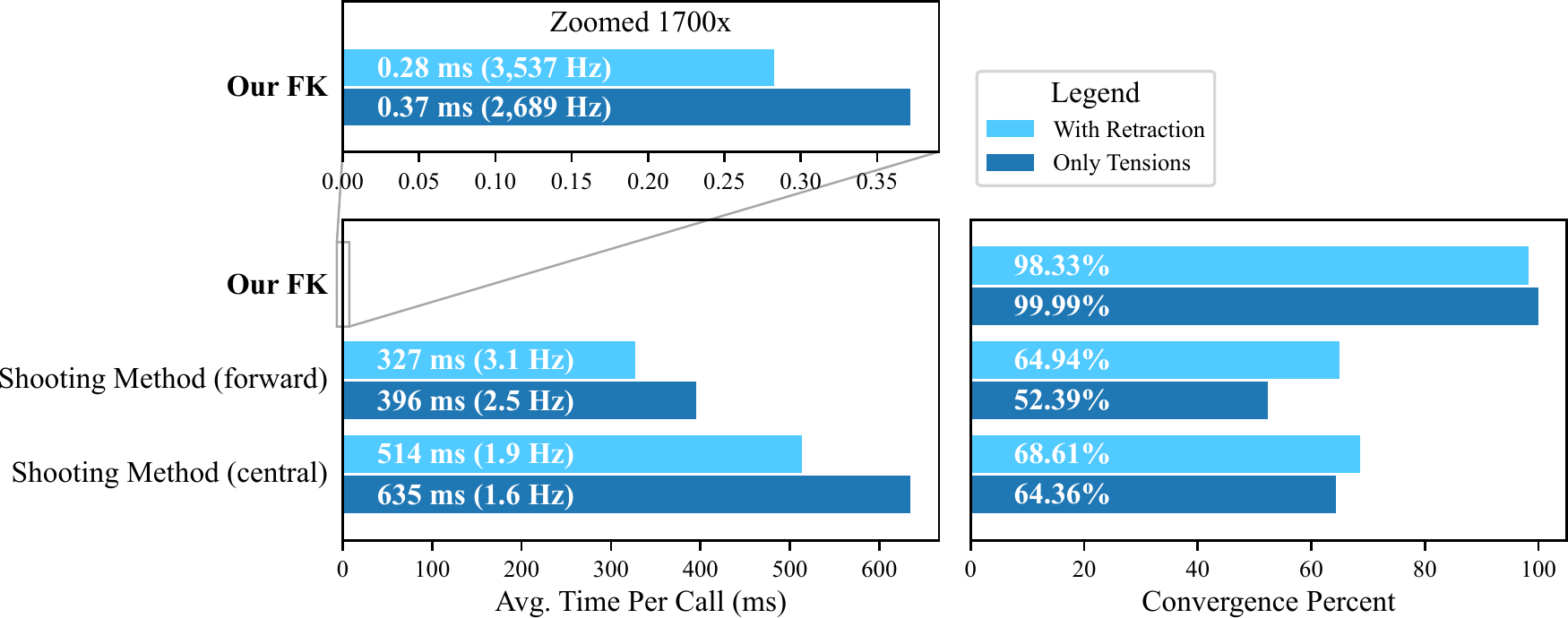}
  \caption{
    Comparison of FK methods on \num{10 000} randomly sampled configurations from $Q$.
    We compare our FK method versus the shooting method using forward-differences and central-differences for computing the Jacobian in the Levenberg-Marquardt algorithm.
    We ran the experiment with retraction enabled and disabled, showing the effect of integrating over a shorter robot length.
    (Left) A comparison of the average time per FK shape computation.
    Above, a zoomed-in view of our FK method's timing shows our method is \num{1100} and \num{1750} times faster on average than the central-differences and forward-differences versions, respectively.
    (Right) The percentage of convergence of FK calls over the \num{10 000} configurations for the shooting method.
    Successful Convergence has a combined force and torque $\ell_2$ residual less than \num{5e-6}.
    Our FK successfully converges on over \SI{98}{\percent} of configurations, whereas the shooting method succeeds on up to \SI{69}{\percent} of configurations.
  }
  \label{fig:fk-implementations}
\end{figure*}

Fig.~\ref{fig:fk-implementations} shows the results of FK on \num{10 000} random configurations from $Q$.
Without retraction enabled, our method performs FK \num{1707} and \num{1064} times faster than the shooting method using central differences and forward differences, respectively.
With the added retraction degree of freedom, our method is \num{1 818} and \num{1157} times faster, respectively.
Our FK method is capable of running at \SI{2.7}{\kHz} without retraction, and \SI{3.5}{\kHz} with retraction on average.

%
%

On the right side of Fig.~\ref{fig:fk-implementations}, the comparison shooting method converges to a residual within the threshold between \SI{52}{\percent} and \SI{69}{\percent} of the time.
Our method achieves convergence for \SI{98.33}{\percent} and \SI{99.99}{\percent} of configurations with and without retraction, respectively.
The inconsistent convergence of the shooting method is likely due to poor initial guesses of $\vec{u}(0)$ and $\vec{v}(0)$ and the optimization getting stuck in local minima.

In all subsequent experiments, non-convergent configurations from the FK computation are considered invalid and removed from $Q_{\text{valid}}$.
We treat them similarly to self-colliding robot configurations.

\subsubsection{Collision-Checking Study}

\begin{figure}
  \centering
  \includegraphics[width=\columnwidth]{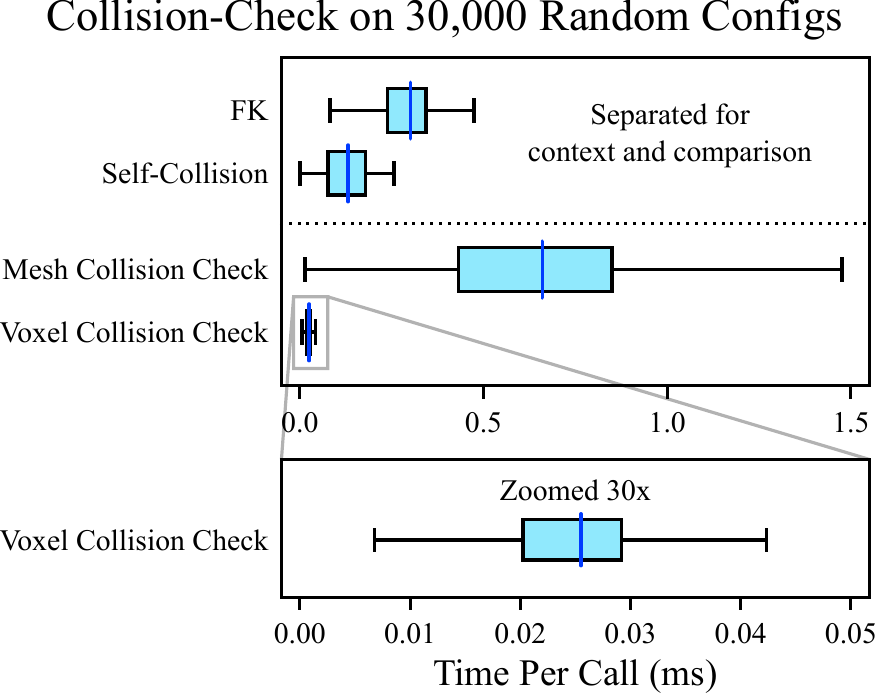}
  \caption{
    Direct timing comparison boxplot between the mesh and voxel collision checkers against the anatomical environment for \num{30 000} random configurations from $Q_{\text{valid}}$.
    Since we run FK and self-collision before doing collision checks against the environment, we separate them out to isolate the collision-checking performance and provide context and comparison.
    On average, the voxel checker executes \num{27.6} times faster than the mesh checker and is now \num{11.8} times faster than our FK and \num{5.2} times faster than the self-collision check.
    In the bottom plot, we show a magnified view of the voxel collision checker's results.
  }
  \label{fig:collision-compare-config}
\end{figure}

To evaluate our voxel-based robot-to-environment collision detection method, we compare our method's performance against the commonly-used strategy of checking bounding capsules along the robot shape against a mesh representation of the anatomical environment.
We refer to our voxel-based method as the voxel checker and the mesh-to-capsule method as the mesh checker.
We implement the mesh checker using the Flexible Collision Library (FCL)~\cite{Pan2012_ICRA}, version 0.5.0.
We compare our voxel checker to the mesh checker in performing collision detection for \num{30 000} configurations randomly sampled from $Q_{\text{valid}}$ (a.k.a., we perform collision checks on only valid configurations and we reject samples in self-collision, with non-converging FK, or outside of tendon length limits).
We sample uniformly for all degrees of freedom, except we use spherical sampling for retraction as seen in Fig.~\ref{fig:config-sampling}.
The results in Fig.~\ref{fig:collision-compare-config} demonstrate that our voxel checker performs collision detection on average \num{27.6} times faster than the mesh checker for single configurations.

Collision checking involves (i) an FK computation, (ii) a self-collision check, and (iii) a collision check against the environment.
This timing measurement separates out the FK and self-collision portions shared between the mesh and voxel collision checkers.
Our voxel checker runs \num{11.8} times faster than our FK and \num{5.2} times faster than the self-collision checker.
Collision detection with our voxel checker now spends \SI{65}{\percent}, \SI{29}{\percent}, and \SI{6}{\percent} of its runtime in FK, self-collision, and environment collision checks, respectively, whereas the mesh checker's runtime is split \SI{26}{\percent}, \SI{12}{\percent}, and \SI{62}{\percent}, respectively.

Voxelization of the robot backbone consumes the vast majority of the time spent in our voxel checker.
The voxelization step takes \SI{98.5}{\percent} of the computation time for voxel collision checking.
This means that for pre-voxelized vertices and edges, collision-checking is approximately \num{65} times faster when not counting FK and self-collision.
However, the precomputation alleviates the need to run FK and self-collision, enabling collision checking against these precomputed vertices and edges \num{1 100} times faster at \SI{3.0}{\MHz} as opposed to the \SI{2.3}{\kHz} otherwise achievable using FK, self-collision, voxelization, and then collision checking.
This motivates our precomputed voxelizations in the Precompute phase and the roadmap pruning in the Load phase.

These results show that our voxel-based collision checking without precomputation is \num{2.5} times faster than the traditional mesh checker.
We also show that our method's collision checking is an additional \num{1 100} times faster with precomputation than without precomputation.

\subsubsection{Dynamic Edge Discretization Study}

\begin{figure}
  \centering
  \includegraphics[width=\columnwidth]{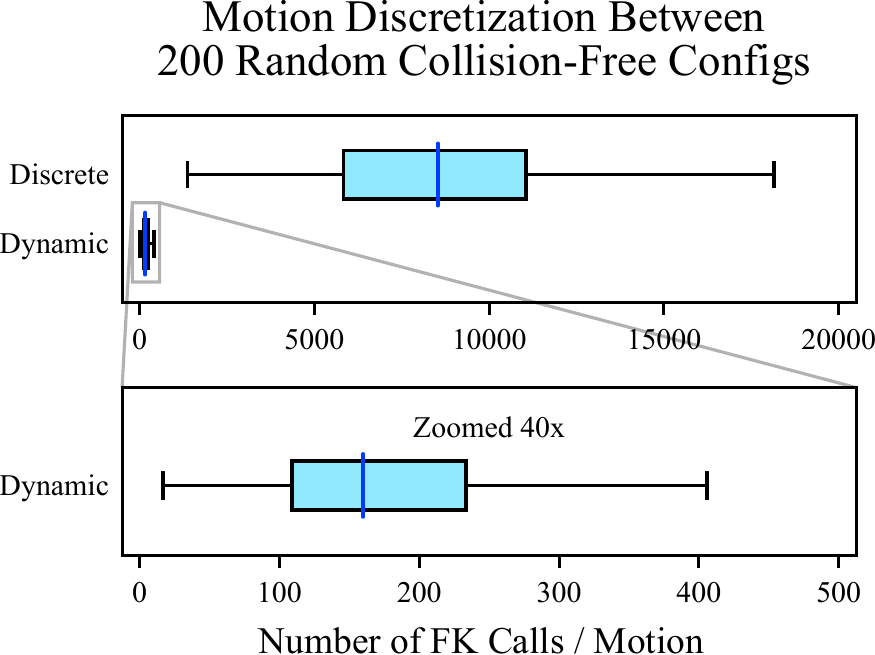}
  \caption{
    Comparison of our dynamic motion discretization compared to a traditional discrete checker between \num{200} randomly selected collision-free configuration pairs.
    Since both use our fast voxel collision checker, we compare the number of discretizations, i.e., the number of forward-kinematic (FK) computations, required for each voxelized swept-volume motion.
    The discrete checker was set to the worst-case discretization of our dynamic checker to ensure we compare for similar safety.
    On average, the dynamic checker takes \num{174} FK calls and the discrete checker takes \num{8 594}, making it \num{49} times more expensive to provide the same level of safety.
    In the bottom plot, we show a magnified view of our dynamic motion discretizer's results.
  }
  \label{fig:collision-compare-edge}
\end{figure}

Beyond the collision detection of single configurations, one of the major costs of motion planning is validating---via collision detection---the motions between configurations, i.e., edges.
Typically, edge validation is done by discretizing the edge into equally-spaced configurations.
We next evaluate our dynamic discretization approach against an equally spaced discrete motion validator.
In Fig.~\ref{fig:collision-compare-edge}, we compare the number of discretizations (i.e., number of FK calls) required in edge validation, since that overwhelmingly determines performance.
As motions are typically only checked if both endpoints are collision-free, we evaluate with only collision-free endpoint configurations, i.e., from $Q_{\text{free}}$.
We see that for \num{200} randomly selected motions, our dynamic edge strategy requires only \num{174} discretizations on average versus \num{8 594} required by the discrete motion validator, allowing it to be \num{49.4} times faster on average.

\subsubsection{Motion Planner Comparison Study}
\label{sec:eval:comparison}

\begin{figure}
  \includegraphics[width=\columnwidth]{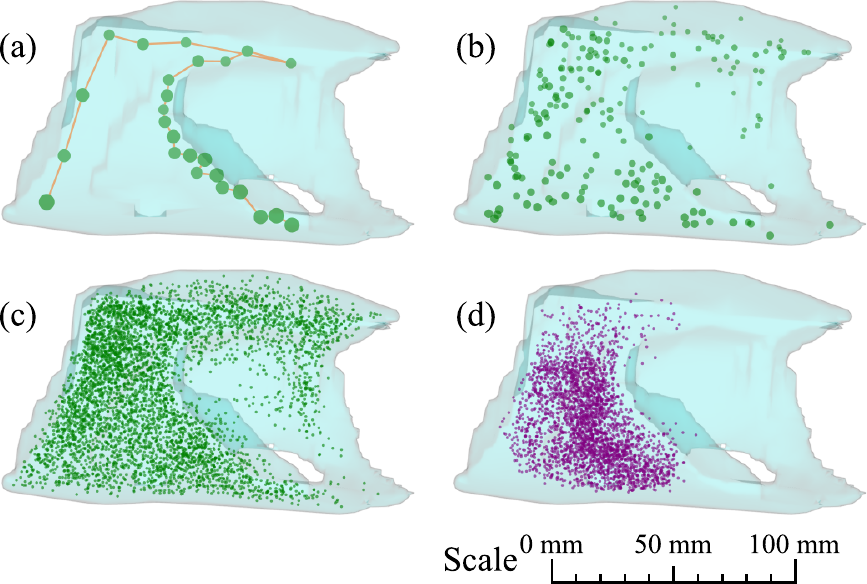}
  \caption{
    Tip positions used in timing experiments within our anatomical environment (environment shown in Fig.~\ref{fig:ct-slices}).
    \textbf{(a)}
    A sequence of \num{25} manually chosen tip positions, largely chosen near the surface of the collapsed lung.
    The collapsed lung is the large obstacle protruding from the right side into the middle of the anatomical environment.
    This sequence represents desired tip positions a clinician may provide in exploring the anatomical environment and is used to compare our method against other off-the-shelf planners.
    The ordering is shown with the orange line starting from the bottom-right of the region.
    \textbf{(b)}
    A sequence of \num{200} randomly chosen tip positions within the anatomy, randomly ordered, and used in the motion-planning comparison study.
    \textbf{(c)}
    The same points from (b) but with an additional \num{4 800} points for a total of \num{5 000} random tip position points, used in the ablation study.
    \textbf{(d)}
    Initial collision-free roadmap tip positions used by our method, shown here for comparison against the evaluation and to qualitatively gauge the evaluation difficulty.
    Many of these tip positions are unreachable by the robot, yet the IK solver and motion planners will attempt to minimize goal tip-position error.
  }
  \label{fig:chosen-lung-points}
\end{figure}

We next compare our full method in a simulated anatomical environment against multiple competing motion planning algorithms that are commonly used in practice.

For our full method, we generated a roadmap with $N_r = \text{\num{30 000}}$ vertices (the same configurations from the collision experiment) followed by \num{570 129} connecting edges (from the PRM$^*$ connection strategy).
We employed 16 concurrent threads in roadmap generation, voxelizing all \num{30 000} vertices in \SI{1.4}{\s}, and voxelizing all \num{570 129} edges in \SI{20}{\minute}, resulting in a \SI{1.6}{\gibi\byte} roadmap file (each occupied \num{4 x 4 x 4} voxel block requires \SI{11}{\byte} on disk).
As a reminder, we perform this roadmap generation without knowledge of any specific anatomical environment.
Fig.~\ref{fig:config-sampling}(b) shows the tip positions of our precomputed roadmap and Figs.~\ref{fig:config-sampling}(d) and~\ref{fig:chosen-lung-points}(d) show the remaining roadmap tip positions after pruning in the Load phase.

There are two popular strategies for motion planning to a workspace end-effector goal, as we are doing in our problem.
In the first, a goal region around the workspace goal point is defined, and any configuration evaluated during planning that places the end-effector in that region is treated as a goal configuration.
In the second, configurations that achieve the workspace goal with the end-effector are generated in advance, typically via IK, and the planning happens in the configuration space to those configuration space goals.
Our method leverages the second strategy, however we evaluate against motion planners that employ both of these conventions, as implemented in the Open Motion-Planning Library (OMPL)~\cite{Sucan2012_RAM}, version 1.5.0.

For the first, we use RRT as a comparison method with a goal region defined by a ball of radius \SI{0.5}{\mm} around the desired workspace goal.
This radius matches the same stopping criteria used for IK in other methods and is slightly less than the \SI{0.59}{\mm} width of a single voxel.
RRT is a single-query motion planner that grows a tree of collision-free configurations until reaching the goal.
In this case, RRT expands its tree in configuration space until either reaching a configuration that places the robot's tip within the goal region, or times out, in which case, it returns the path to the configuration with its tip closest to the workspace goal.

The second strategy, utilizing computed goal configurations from the specified workspace goal, enables efficient approaches such as goal biasing and bidirectional search.
For this strategy, we compare against three popular planners, RRTConnect, PRM, and LazyPRM, that are given a goal configuration calculated using IK.
RRTConnect is a variant of RRT that utilizes bidirectional search, attempting to connect trees growing from both the start and goal configurations.
PRM is a traditional multi-query motion planner, reusing the same roadmap in future queries.
LazyPRM adds lazy evaluation to PRM, only collision-checking vertices and edges after finding a candidate path.
These planners do not use our \RoadmapIK algorithm (Alg.~\ref{alg:roadmap-ik}), but rather a more traditional IK algorithm, which we label \NormalIK in this work (see Alg.~\ref{alg:normal-ik} in Appendix~\ref{apdx:ik}).
\NormalIK uses the same LM optimizer for IK as our \RoadmapIK, but instead of finding the closest connection to the roadmap, if the IK solution is in collision, then we project it into $Q_{\text{free}}$ by taking small steps backwards towards the initial guess until finding a collision-free configuration.
The initial guesses we employ for \NormalIK start with the current robot configuration (i.e., the end of the previous plan), but allowing up to $k_{\text{restarts}}$ random restarts with random samples from $Q_{\text{free}}$ if the IK solution projected into $Q_{\text{free}}$ is not within the specified tolerance, ultimately returning the closest one if we use all $k_{\text{restarts}}$ restarts.

We compare our method's performance against the above motion planners on two datasets.
The first dataset is a sequence of \num{25} goal tip positions along the outer wall of the collapsed lung on the right of the anatomical region, as seen in Fig.~\ref{fig:chosen-lung-points}(a).
We manually chose this goal tip sequence to represent potential areas that a clinician may wish to explore, specifically visiting nearby points to explore the outer surface of the lung and pleural space.
On this first dataset, we give a timeout of \SI{120}{\s}.
This timeout is much higher than we would accept for an interactive system, but this is to enable the competing planners to achieve as high of accuracy as possible.
We only apply this timeout to planning after performing IK.
We task the methods with reaching each of these goal tip positions in sequence, simulating an interactive, supervisory-control scenario in which these goals are desired one after another.

The second dataset consists of \num{200} randomly chosen tip positions within the anatomy as seen in Fig.~\ref{fig:chosen-lung-points}(b).
For this dataset, we use a timeout of \SI{20}{\s}, which is still higher than ideal for an interactive system, but lower than the \SI{120}{\s} used on the first dataset.
This timeout is intended to highlight the best these planners can do on tip accuracy given the more limited (but still large from an interactive sense) amount of time.
As above, we task the methods with reaching each of these goal tip positions in sequence, one after another.

In this comparison study, all planners utilize our FK model, our voxel collision checker, and our dynamic edge discretization method.
We set the discrete sampler distance thresholds to the same thresholds utilized by the dynamic edge discretization study.
This allows us to evaluate the contribution of our motion-planning method separately from our improvements to kinematic modeling and collision checking.
%
%
Our method's \RoadmapIK uses up to $k_{\text{IK}} = \num{5}$ nearest neighbors for IK if it does not converge to the goal within the tip-error threshold.
\NormalIK uses up to $k_{\text{restarts}} = \num{10}$ and $k_{\text{restarts}} = \num{2}$ random restarts on the \num{25} and \num{200} goal sets, respectively.
Additional implementation details for \RoadmapIK and \NormalIK are provided in Appendix~\ref{apdx:ik}.

\begin{figure*}
  \centering
  \includegraphics[width=\textwidth]{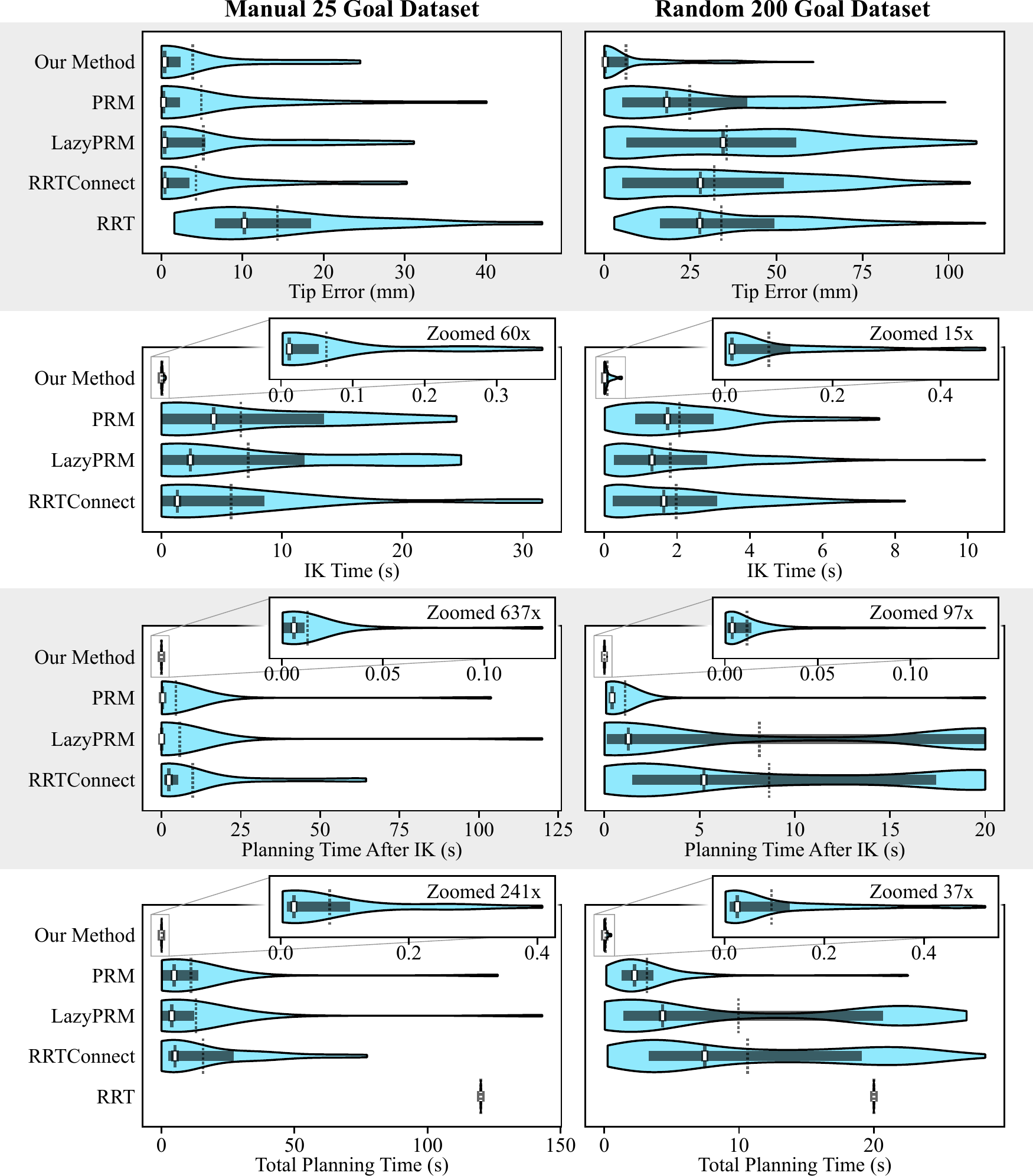}
  \caption{
    Planner comparison for
    \textbf{(left column)} \num{25} interior workspace goals from Fig.~\ref{fig:chosen-lung-points}(a) with a timeout of \SI{120}{\s} and up to $k_{\text{restarts}} = \text{\num{10}}$ random restarts for \NormalIK, and
    \textbf{(right column)} \num{200} random interior workspace goals from Fig.~\ref{fig:chosen-lung-points}(c) with a timeout of \SI{20}{\s} and up to $k_{\text{restarts}} = \text{\num{2}}$ random restarts for \NormalIK.
    These violin plots show the density curves of measured values based on approximate frequency.
    Overlaid on top, we show the median and quartiles as a cross with a white center and the mean as a dotted line.
    \textbf{Row 1}
    compares the returned plan's endpoint tip error compared to the goal position.
    \textbf{Row 2}
    compares the time to run IK to the given goal position before generating a plan.
    Our method uses the \RoadmapIK algorithm with $k_{\text{IK}} = 5$ on both goal sets.
    PRM, LazyPRM, and RRTConnect use the \NormalIK algorithm.
    \textbf{Row 3}
    compares the time to generate a plan after IK.
    We truncate this timing to the timeout value.
    \textbf{Row 4}
    compares total time for planning starting from when the goal position is given until a plan is returned (sum of rows 2 and 3).
    For rows 2--4, we display a magnified view of our method's results.
    The RRT planner timed out for every goal in both goal sets, returning the best found plan.
    RRT uses a workspace goal region instead of IK, so it is excluded from the comparison of IK and plan generation time (rows 2 and 3), but is included in the comparison of total planning time (row 4) after truncating to the timeout.
  }
  \label{fig:comparison-25-and-200}
\end{figure*}

The results can be seen in the violin plots of Fig.~\ref{fig:comparison-25-and-200} for the \num{25} manually-selected goal tip positions in the left column and the \num{200} randomly generated goal tip positions in the right column.
Below, we compare and discuss tip error and timing on average (the dotted lines in Fig.~\ref{fig:comparison-25-and-200}) and in the worst-case (max value of each violin in Fig.~\ref{fig:comparison-25-and-200}).
However Fig.~\ref{fig:comparison-25-and-200} also shows min, median, and quartile information.

We first discuss a comparison of tip error (Fig.~\ref{fig:comparison-25-and-200}, first row).
Our method,
PRM,
LazyPRM,
RRTConnect,
and RRT
on the \num{25} goals achieve an average tip error of
\SI{3.9}{\mm},
\SI{4.9}{\mm},
\SI{5.2}{\mm},
\SI{4.3}{\mm}, and
\SI{14.3}{\mm}
per goal, respectively, and for the \num{200} goals achieve an average tip error of
\SI{6.3}{\mm},
\SI{24.8}{\mm},
\SI{35.5}{\mm},
\SI{31.9}{\mm}, and
\SI{34.0}{\mm}
per goal, respectively.
Our method achieves \SI{9}{\percent} and \SI{75}{\percent} lower average tip error than the best competitors (LazyPRM and PRM) on the \num{25} and \num{200} goal sets, respectively.
In the worst-case,
our full method,
PRM,
LazyPRM,
RRTConnect, and
RRT
on the \num{25} goals have tip error of
\SI{24.5}{\mm},
\SI{40.1}{\mm},
\SI{31.1}{\mm},
\SI{30.3}{\mm}, and
\SI{46.9}{\mm}, respectively,
and on the \num{200} goals have tip error of
\SI{60.8}{\mm},
\SI{99.1}{\mm},
\SI{108}{\mm},
\SI{106}{\mm}, and
\SI{111}{\mm}, respectively.
Our method achieves \SI{19}{\percent} and \SI{39}{\percent} lower worst-case tip error than the best competitors (LazyPRM and RRT) on the \num{25} and \num{200} goal sets, respectively.

We recognize that in OMPL's implementation, PRM and LazyPRM often return an empty plan after timing out.
Also, after timing out, OMPL's RRTConnect returns a partial plan, which sometimes begins from the goal configuration instead of from the start, which we detect and replace with an empty plan.
Besides RRT (which timed out for every goal in both goal sets), PRM, LazyPRM, and RRTConnect had \num{0}, \num{1}, and \num{0} timeouts on the \num{25} goal set, respectively, and on the \num{200} goal set, had \num{5}, \num{69}, and \num{44} timeouts, respectively.
LazyPRM always returned an empty plan upon timeout, but PRM and RRTConnect returned \num{0} and \num{8} empty plans for their \num{5} and \num{44} timeouts on the \num{200} goal set, respectively.
If we omit goals for which an empty plan was returned, then
LazyPRM decreases to an average tip error of \SI{4.54}{\mm} on the \num{25} goal set,
and 
PRM,
LazyPRM, and
RRTConnect
decrease to an average tip error of
\SI{23.2}{\mm},
\SI{31.1}{\mm}, and
\SI{33.5}{\mm}
on the \num{200} goal set,
respectively.
Our method still achieves \SI{9}{\percent} and \SI{73}{\percent} lower tip error than the best competitors, RRTConnect and LazyPRM, on the \num{25} and \num{200} goal sets, respectively.
Only LazyPRM's worst-case tip error decreases by omitting empty plans to \SI{94.0}{\mm} on only the \num{200} goal set, which decreases our improvement to \SI{35}{\percent} lower tip error.

This high tip error on the \num{200} random goal set is not surprising since many requested tip positions are not within the reachable workspace of the simulated tendon-driven robot, as seen by comparing Fig.~\ref{fig:chosen-lung-points}(b) and~\ref{fig:chosen-lung-points}(d).
Instead we emphasize our method's significantly lower tip error on average and in the worst-case.

Next, we compare planning times.
We separate the total planning time for each goal (Fig.~\ref{fig:comparison-25-and-200}, fourth row) into IK time and planning time (Fig.~\ref{fig:comparison-25-and-200}, second and third rows, respectively).
Since RRT uses goal regions instead of IK, it is omitted from the second and third rows of Fig.~\ref{fig:comparison-25-and-200}.
We use the IK timing in the second row of Fig.~\ref{fig:comparison-25-and-200} to evaluate our method's \RoadmapIK algorithm versus \NormalIK which is used by PRM, LazyPRM and RRTConnect.
Our method,
PRM,
LazyPRM, and
RRTConnect
solve IK on average in
\SI{0.063}{\s},
\SI{6.6}{\s},
\SI{7.2}{\s}, and
\SI{5.8}{\s}
on the \num{25} goal set, respectively, and on the \num{200} goal set, they average
\SI{0.082}{\s},
\SI{2.6}{\s},
\SI{1.8}{\s}, and
\SI{2.0}{\s}, respectively.
Our method's IK (\RoadmapIK) is \num{91} and \num{22} times faster on average, and in the worst-case is \num{67} and \num{16} times faster for the \num{25} and \num{200} goal sets, respectively (against LazyPRM and PRM, respectively, both employing \NormalIK).

For this interactive system, total planning time is an important metric as it indicates the time the clinician must wait before specifying a new goal tip position.
The average total time per goal for
our method,
PRM,
LazyPRM,
RRTConnect,
and RRT
on the \num{25} goals were
\SI{0.077}{\s},
\SI{11.1}{\s},
\SI{12.9}{\s},
\SI{15.6}{\s},
and \SI{120}{\s}, respectively,
and on the \num{200} goals were
\SI{0.094}{\s},
\SI{3.2}{\s},
\SI{10.0}{\s},
\SI{10.6}{\s},
and \SI{20}{\s}, respectively.
The RRT planner for both datasets reached the timeout for every goal position in both datasets (timeout of \SI{120}{\s} and \SI{20}{\s} for the \num{25} and \num{200} goals, respectively).
Our full method is \num{145} times and \num{33} times faster on average than the best competing planner, PRM, for the \num{25} and \num{200} goals, respectively.
Importantly, all compared planners, including PRM, have much higher worst-case timing than our method.
The slowest plans for
our method,
PRM,
LazyPRM,
RRTConnect,
and RRT
on the \num{25} goals were
\SI{0.41}{\s},
\SI{126}{\s},
\SI{143}{\s},
\SI{77}{\s},
and \SI{120}{\s}, respectively,
and on the \num{200} goals were
\SI{0.52}{\s},
\SI{22.6}{\s},
\SI{26.9}{\s},
\SI{28.3}{\s},
and \SI{20}{\s}, respectively.
For the competing planners, PRM had the fastest worst-case timing on the \num{25} goals.
RRT performed the best in the worst-case on the \num{200} goals only because the timeout of \SI{20}{\s} was only applied to planning after performing IK, which this RRT planner does not employ.
As previously mentioned, LazyPRM had \num{1} timeout, and on the \num{200} goals, PRM, LazyPRM, RRTConnect, and RRT had \num{5}, \num{69}, \num{44}, and \num{200} timeouts, respectively.
Compared against RRTConnect on the \num{25} goals and RRT on the \num{200} goals, our method's worst-case timing was \num{189} and \num{38} times faster, respectively.
Against our definition of less than one second for interactive-rate planning, our method's worst-case timing of \SI{0.523}{\s} is well within that threshold, meaning our method achieves interactive-rate planning even in the worst case.

\subsubsection{Full Contributions Comparison Study}

\begin{figure}
  \centering
  \includegraphics[width=\columnwidth]{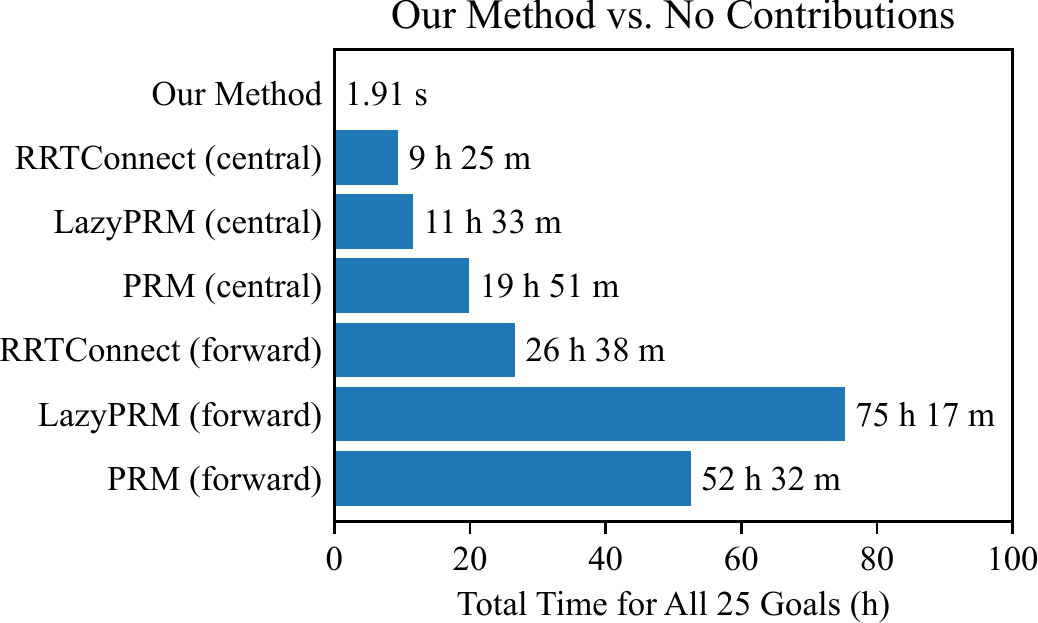}
  \caption{
    Planner comparison for the \num{25} interior workspace goals from Fig.~\ref{fig:chosen-lung-points}(a) with a timeout of \SI{16}{\hour} and up to $k_{\text{restarts}} = \text{\num{10}}$ random restarts for \NormalIK.
    This comparison has RRTConnect, LazyPRM, and PRM utilize the mesh collision checker, the equally-spaced edge evaluator, and the slow FK using the shooting method with forward differences or central difference, marked with (forward) or (central), respectively.
    We compare against our full method to plan to all \num{25} interior workspace goals.
    Timeouts occurred only with forward differences for RRTConnect (\num{1} timeout), LazyPRM (\num{4} timeouts), and PRM (\num{1} timeout).
    With a timeout of \SI{16}{\hour}, the reported planning time for planners using forward-differences is effectively a lower bound.
  }
  \label{fig:slow-comparison-25}
\end{figure}

The prior study evaluates our method against competing off-the-shelf motion planners that are given three of our contributions: (i) our fast FK model, (ii) our fast voxel collision method, and (iii) our dynamic motion discretizer.
These contributions were evaluated separately above.
However, in order to evaluate all of our contributions together, we perform another comparison against competing motion planners that do not employ any of our contributions.
We evaluate RRTConnect, PRM, and LazyPRM on the \num{25} goals dataset with the same settings mentioned previously.
In the competing methods we utilize the traditional mesh collision checker without our dynamic edge evaluator, and we use the slow shooting-method FK kinematic model with forward-differences and central-differences.
We set a large timeout of \SI{16}{\hour} per goal for the competing methods without our contributions.
In Fig.~\ref{fig:slow-comparison-25}, we show the timing for planning to all \num{25} goals.
The best competing planner was RRTConnect with central differences, which took \SI{9}{\hour} \SI{25}{\minute} to generate plans for a sequence of \num{25} goals.
Here, Central differences outperforms forward differences because of the higher convergence, despite the higher performance cost in calculating the numerical Jacobian.
In comparison, our method is \num{17 700} times faster, planning to the same sequence of \num{25} goals in \SI{1.91}{\s}.

\subsubsection{Ablation Study}

We next perform an ablation study in the simulated anatomical environment, in which we enable and disable various aspects of our method, demonstrating the contribution of each aspect of our full method.
The results are seen in Table~\ref{tbl:ablation-5000}.

\newcommand{\narrowdoublestack}[2]{\shortstack{#1\\#2}}

\newcommand{\BN}[2]{\textbf{\tablenum[table-format = \1]{\2}}}

\begin{table*}
  \centering
  \caption{
    Ablation Study for a random sequence of \num{5 000} random interior workspace goals (see Fig.~\ref{fig:chosen-lung-points}(b)).
    The ablation baseline is equivalent to our full method with lazy collision checks and without pre-voxelization.
    Each row adds one more component of our method, starting with adding in pre-voxelization, then pre-checking collisions at load time.
    For loading the roadmap from disk with a single core, we show the load time and the peak computer memory used in the first two columns.
    The \# Failures column shows how many planning queries failed to generate collision-free paths to the \RoadmapIK solution of the \num{5 000} goals.
    This demonstrates the effectiveness of our full method's guarantee of a collision-free path to the IK solution by ensuring the roadmap consists of a single connected component.
  }
  \label{tbl:ablation-5000}
  \begin{tabular}{
      l |
      S [table-format = 2.2]
      S [table-format = 5]
      |
      S [table-format = 3]
      S [table-format = 1.3(3)]
      S [table-format = 1.3(3)]
      S [table-format = 1.3(3)]
      |
      S [table-format = 1.1(3)]
    }
    \toprule
    {}
      & {\narrowdoublestack{Load Time}{(\si{\s})}\hspace{-.3em}}
      & {\hspace{-.3em}\narrowdoublestack{Load Mem}{(\si{\mebi\byte})}}
      & {\narrowdoublestack{\# Failures}{(of \num{5 000})}\hspace{-.5em}}
      & {\narrowdoublestack{IK Time}{(\si{\s})}}
      & {\narrowdoublestack{Plan Time}{(\si{\s})}}
      & {\narrowdoublestack{Time/Goal}{(\si{\s})}}
      & {\narrowdoublestack{Tip Error}{(\si{\mm})}}
      \\
\midrule
  {LazyPRM + pregen graph + \RoadmapIK}     &  1.03 & \B 166 &  177 &    0.064 +- 0.125 &    0.039 +- 0.325 &    0.104 +- 0.351 &    5.8 +- 14.6 \\
  {+ pre-voxelized vertices}                &  1.22 &    348 &  175 &    0.064 +- 0.123 &    0.038 +- 0.312 &    0.103 +- 0.339 &    5.8 +- 14.5 \\
  {+ pre-voxelized edges}                   & 10.47 &  10369 &  176 &    0.065 +- 0.126 &    0.029 +- 0.326 &    0.095 +- 0.353 &    5.8 +- 14.5 \\
  {+ prune vertices on load}              & \B 0.75 &    454 &  102 & \B 0.058 +- 0.114 &    0.006 +- 0.010 & \B 0.065 +- 0.116 &    5.1 +- 12.8 \\
  {+ check edges on load (our full method)} &  0.79 &    454 & \B 0 &    0.061 +- 0.128 & \B 0.006 +- 0.008 &    0.068 +- 0.128 & \B 4.1 +-  9.1 \\
    \bottomrule
  \end{tabular}
\end{table*}

For this study, we randomly sampled a sequence of \num{5 000} interior goal tip positions within the anatomical environment as seen in Fig.~\ref{fig:chosen-lung-points}(c).
The IK parameters match those of our method in the comparison study.

\begin{figure}
  \centering
  \includegraphics[width=\columnwidth]{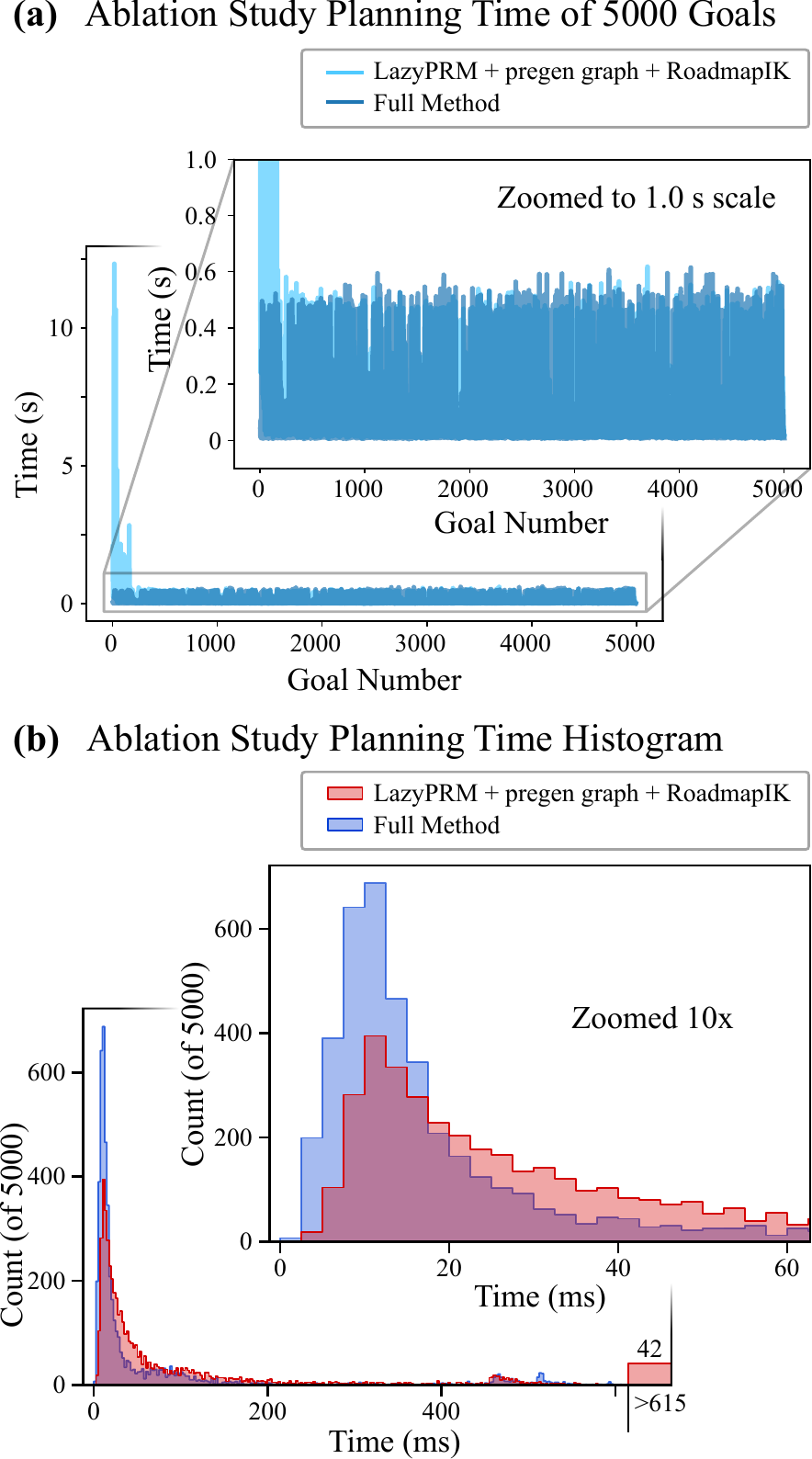}
  \caption{
    We compare two versions of our method, our full method and the ablation baseline, which is equivalent to our full method but using lazy evaluation and a roadmap without voxel precomputation.
    We evaluate the time required to generate a plan starting from when the desired tip position is given on a sequence of \num{5 000} random interior goal points in the anatomical environment (see Fig.~\ref{fig:chosen-lung-points}(b)).
    \textbf{(a)} The timing on each goal point between the two methods and a magnified section showing times for runs below one second, with the ablation baseline shown in light blue and our full method shown in dark blue.
    The baseline method exhibits extreme slowdown on early goals with a maximum time of \SI{12.3}{\s} but improves with time as the roadmap is lazily pruned.
    In contrast, our full method demonstrates high efficiency throughout the entire experiment with a maximum time of \SI{0.615}{\s}.
    \textbf{(b)} Timing histograms in milliseconds between zero and our full method's worst-case timing of \SI{615}{\ms} (with the ablation baseline having \num{42} timings above this threshold), and a section magnified by a factor of ten, with the ablation baseline shown in transparent red and the full method shown in transparent blue.
    Each bin represents a range of \SI{2.5}{\ms}.
    The baseline method achieves a similar mode to our full method, but with a much larger and longer tail towards higher timings.
    }
  \label{fig:milestone-timing-compare-lineplot}
\end{figure}

The base level of this ablation study is equivalent to LazyPRM with a pregenerated roadmap of configurations (LazyPRM + pregen graph + \RoadmapIK in Table~\ref{tbl:ablation-5000}).
We add to this our contributions of pre-voxelized vertices, pre-voxelized edges, collision-detecting the vertices when loading, and collision-detecting the edges when loading, resulting in our full method.
In addition to the results presented in Table~\ref{tbl:ablation-5000}, we present timing comparisons between the base level version of our method and our full method in Fig.~\ref{fig:milestone-timing-compare-lineplot}.

For these \num{5 000} points, the base implementation performed quite well on average, able to replan at \SI{9.6}{\Hz} on average, whereas our full method achieves \SI{14.8}{\Hz} on average.
However, as seen in Fig.~\ref{fig:milestone-timing-compare-lineplot}, the base implementation starts out solving the first few motion plans in upwards of \SI{12.3}{\s} each.
This worst-case timing precludes interactive applications such as surgical procedures.
In contrast, our full method is able to maintain a worst-case planning time of \SI{0.61}{\s}.
%
%
The baseline method and our full method generate plans in less than \SI{20}{\ms} for \SI{33}{\percent} and \SI{59}{\percent} of the goals, respectively.

Adding cached voxelization of vertices and edges increases the loading time required by the method.
The roadmap without any cache utilizes \SI{11}{\mebi\byte} of disk space.
Adding voxelized vertices and edges increases that file size to \SI{30}{\mebi\byte} and \SI{1601}{\mebi\byte}, respectively.
Our roadmap file is a custom binary serialization of the graph and voxel objects.
Loading the full graph into memory (Table~\ref{tbl:ablation-5000} row 3) takes \SI{10.1}{\gibi\byte} of memory and \SI{10.5}{\s} of load time.
The load time is a single, one-time cost before performing motion planning, and is therefore acceptable for our problem.

When pruning in-collision vertices and edges while reading the roadmap from file, the peak memory required during loading of the roadmap drops from \SI{10.1}{\gibi\byte} to less than \SI{0.44}{\gibi\byte}.
Also, unsurprisingly, the performance improves drastically as we are now working with a fully-connected collision-free graph that requires no removals, and the first found solution will contain only collision-free vertices.
However, there were a few surprising results as a consequence of removing laziness from vertex and edge collisions by pruning at loading-time.
It is surprising that loading a large roadmap containing voxelized vertices and edges while performing collision checks loads faster than a roadmap that contains no cached voxel sets.
We found that constructing the full graph in memory when not pruning accounts for the additional loading time.
When performing collision checks during loading, we only construct the portion of the graph that is collision-free.
After removing in-collision vertices and edges, and removing vertices disconnected from the main component, the roadmap only contains \num{2 833} vertices and \num{21 839} edges, down from \num{30 000} vertices and \num{570 129} edges, causing the generated graph to be much smaller and less costly to generate.

As we do not sample random points during planning, our method is completely deterministic after loading our pregenerated roadmap.
The difference in tip error in the ablation study is due to the number of times we fail to generate a collision-free path through the roadmap.
This is because the added IK solution sometimes gets added to a portion of the graph that is unreachable from the main connected component.
We see the number of failures drop significantly down to \num{102} from \num{176} when we prune in-collision vertices and remove disconnected components when loading the roadmap.
This number further drops to zero when we further prune in-collision edges and remove disconnected components when loading the roadmap.
This demonstrates the effectiveness of our full method's guarantee by construction of a collision-free path to the IK solution.

%
%

The tip error achieved by our full method for the \num{5000} points is \SI{4.1}{\mm} on average.
We note that these requested goal positions were sampled uniformly at random within the anatomical region.
Comparing Fig.~\ref{fig:chosen-lung-points}(c) with our initial roadmap in Fig.~\ref{fig:chosen-lung-points}(d), we can see qualitatively that many of the points are highly challenging to reach from our roadmap.
We further have no guarantees that these sampled tip goal positions are within the reachable workspace of the robot.
However, \SI{80.2}{\percent} of the plans from our full method were within the \SI{0.5}{\mm} stopping criteria for \RoadmapIK.

\section{Limitations of the Study and Future Work}
\label{sec:limitations}

%
%

Although we significantly reduce computation time of tendon robot shape computations and collision checks, we recognize that self-collision checks consume \SI{29}{\percent} of collision-checking runtime, which may be further reduced using, e.g., a hierarchical approach as described in~\cite{Franco2014_IJCSI}.
We leave as future work implementation improvements to self-collision, and developing a principled approach for determining adequate discretizations for robot backbone spacing (for both self-collision and voxelization).

%
%

Our implementation of the compared shooting method FK model, although standard, could potentially be improved in a number of ways, including better initial guesses to LM, parameter-space scaling, and incremental changes to tension values for improved convergence. 
However it is outside of the scope of this work to augment such a method with these improvements.
Further, our fast FK model solves for $\vec{u}(0)$ and $\vec{v}(0)$ in the case of no external loads, and we could potentially utilize this updated guess in the shooting method when there are external loads.
Future work to improve the FK model in the presence of external forces and torques would enable motion planning and model-based control for tasks which require the application of force to the surrounding environment.

%
%

In the Precompute phase, we voxelize the robot shape and swept volume motions already in the insertion orientation and in the medical image space's.
However, in practice, we would not have the insertion position and orientation for voxelizing before being given the anatomical environment.
A translational offset is trivial to handle, but a different insertion orientation would require resampling to ensure all voxel sets have matching voxel discretizations and frames.
This may be achieved by either resampling the image into the robot voxel space, resampling the pre-voxelized robot shapes into the medical image space, or by finding another solution besides pre-voxelization.
We leave as future work investigating alternative precomputation shape representations that are easily convertible to voxels after being given the orientation, in order to maintain the ability to precompute once for each robot design and reuse it in many anatomical environments.

%
%

In the comparison and ablation studies, we evaluated the effectiveness of using arbitrary interior tip positions in the anatomical region without considering the reachable workspace of the robot.
Although this approach is inspired by the application of controlling the robot to explore the anatomical region, it is not clear how closely the \RoadmapIK approach is able to get to difficult-to-reach tip positions that are attainable.
In future work, we wish to consider workspace reachability metrics within the constrained anatomical environment.

%
%

Our current problem formulation only considers finding a collision-free trajectory to the goal tip position and optimizing for proximity of the robot's tip at the end of the path to the goal.
However, it does not consider any characteristics of the motion the robot uses to get there, other than collision avoidance.
In future work it would be beneficial to incorporate some notion of optimality to choose the best desired path among all possible paths, under additional objective functions.

%
%

Our approach presents a supervisory-control method intended to be used by a clinician to explore an anatomical region.
This work could be strengthened by including user studies, preferably with clinicians, using the system in simulation or on a physical robot.
This is certainly the next step toward making this a viable system used in surgical settings.
Also, while waiting for the next goal tip position from a clinician, the planner could actively collision-check unevaluated lazy edges and add to its roadmap either randomly or by utilizing IK to fill in missing regions within the anatomy.
In this way, we could perform repairs to the roadmap to connect the disconnected components rather than removing them during the Load phase.

%
%

We recognize that surgeons will likely require the robot to face a particular region in the anatomy rather than simply having the tip reach a desired location.
This requires inverse kinematic solving of not just position, but also orientation, and may be encompassed more broadly with task planning.
We view this as a particularly important and promising next step for this work, and note that we could augment our approach with interleaved optimization and planning~\cite{Kuntz2019_IROS,Kuntz2020_ISRR} to achieve goal orientations.

%
%

The studies in this work were performed in simulation.
We next intend to reproduce the results on a physical robot, such as is shown in Fig.~\ref{fig:tendon-robot}.

%
%

Finally, in this study, we assume we are planning in a static anatomical environment, which will not be true in most surgical settings.
Because collision checks of pre-voxelized shapes are extremely fast at about \SI{4}{\MHz}, by keeping the voxelized vertices and edges in memory, the list of valid vertices and edges may be updated quickly against a changing environment.
This requires sensing and dynamic updates to the voxelized environment and quick replanning.
In the future, we believe our fast motion planner and collision checker will enable this type of approach with frequent replanning.

\section{Conclusion}
\label{sec:conclusion}

Continuum robots, such as the tendon-driven robot we present in this work, have the potential to increase capabilities in minimally-invasive surgeries.
The increased flexibility and inherent compliance enable access to difficult-to-reach anatomical regions and may improve patient outcomes.

In this work, we present a full motion-planning system for tendon-driven robots capable of being controlled at an interactive-rate.
This is accomplished despite complex kinematic modeling and a highly constrained example anatomical environment via a number of contributions.
We first presented an improved kinematic model for the case of no external loads which, when compared against current modeling approaches, enables approximately \num{1 100} times faster speeds and a much higher convergence success rate of \SI{98.3}{\percent}.
Next, we develop a fast voxel collision method using Octrees and bitwise operations to leverage representations naturally generated from medical imaging, enabling \num{27.6} times faster collision checking than checking bounding capsules along the robot shape against a mesh representation of the anatomical environment.
This sparse voxel representation provides an efficient storage representation for robot shapes and swept volumes between configurations.
This representation empowers us to precompute a large roadmap in the Precompute phase with voxelized vertices and edges that can be checked in \SI{0.8}{\s} during the Load phase.
Finally, during the Supervisory Control phase, we utilize our roadmap with our \RoadmapIK algorithm to achieve high tip accuracy and fast planning, running on average at \SI{14.8}{\Hz}, approximately \num{17 700} times faster than the best evaluated off-the-shelf planner without our contributions.

\appendices
\section{Spherical Sampling}
\label{apdx:spherical-sampling}

For a sphere of radius $\ell$, the likelihood of generating a uniformly random sample within the sphere with a distance $\alpha$ from the center is proportional to the area of points that have that same distance
$
  P(\alpha) \propto \alpha^2
$.
By integration, the Cumulative Distribution Function (CDF) is
$
  C(\alpha) = A \alpha^3 + B
$.
Using $C(0) = 0$ and $C(\ell) = 1$, this simplifies to $C(\alpha) = (\alpha/\ell)^3$ with inverse
$C^{-1}(u) = \ell \sqrt[3]{u}$.
We sample from this distribution using \emph{inverse transform sampling}, which creates a sample from an arbitrary distribution by passing a uniform random variable on $[0, 1]$ into the inverse CDF of that arbitrary distribution.

\section{Forward Kinematic Implementation}
\label{apdx:fk}

Our FK method uses the fixed-point iteration strategy described in Section~\ref{sec:method:fk} to solve for $\vec{v}(0)$ and $\vec{u}(0)$ before performing integration.
We set the maximum number of fixed-point iterations to \num{1000}.
We define the force and torque combined residual as
$
  \sqrt{
    \|\vec{F}_{e,0}\|^2
    +
    \|\vec{L}_{e,0}\|^2
  }
$,
where $\vec{F}_{e,0}$ is the external point force and $\vec{L}_{e,0}$ is the external point torque at the robot's base.
We calculate this with SI units of \si{\newton} for force and \si{\newton\m} for torque.
We set the stopping criteria and define successful convergence by a residual less than \num{5e-6}.

After solving for $\vec{v}(0)$ and $\vec{u}(0)$ from the fixed-point iteration, we integrate~\eqref{eq:uv} with a constant backbone discretization step size of \SI{0.59}{\mm}, the same as the width of a single voxel, resulting in \num{203} discretizations along the backbone when fully extended.
We integrate forward using the Runge-Kutta~4th order method from the Boost Odeint library~\cite{Ahnert2011_AIP} (version 1.65).

The shooting method used as the comparison in our experimental evaluation instead solves~\eqref{eqn:uv-root-finding} using LM as implemented by the levmar library~\cite{Lourakis2004_code}.
We set the maximum number of LM iterations to \num{500} and the finite-difference distance to \SI{1e-8}{\newton}.
We initialize the LM damping factor $\lambda = \num{1e-3}$.
The shooting method uses the same stopping criteria and definition of successful convergence as our method's fixed-point iteration, except the residual is computed for the forces and torques at the tip instead of the base.
These parameters were tuned for speedy convergence on most configurations.

We implement retraction by changing the starting point of integration.
We apply the rotation configuration value after computing the robot shape by a rotation about the $z$-axis.

\section{Dynamic Edge Discretizer}
\label{apdx:edge-discretizer}
\label{apdx:edge-voxelization}

\begin{algorithm}[ht]
  \caption{\VoxelizeFreeEdge}
  \label{alg:voxelize-free-edge}

  \DontPrintSemicolon
  \SetArgSty{} 
  \SetDataSty{} 
  \SetKwInOut{Input}{input}
  \SetKwInOut{Output}{output}
  \SetKwComment{tcp}{\textbf{//} }{}
  \newcommand{\myAlgCommentSty}[1]{\color{blueishgreen}#1}
  \SetCommentSty{myAlgCommentSty}

  \SetKw{KwOr}{or}
  \SetKwData{threshold}{threshold}
  \SetKwData{frontier}{frontier}
  \SetKwProg{myfunc}{Function}{}{}

  \KwIn{ \\
    \quad $\qbegin$: from $Q_{\text{free}}$, voxelize from this \\
    \quad $\qend$: from $Q$, voxelize toward this \\
    \quad $\delta$: discretization threshold distance
  }
  \KwOut{ \\
    \quad $\qreached$: last config in $Q_{\text{free}}$
               from $\qbegin$ toward $\qend$ \\
    \quad $D$: voxelized edge
               from $\qbegin$ to $\qreached$
  }

  \myfunc{$\VoxelizeFreeEdge(
    \qbegin,
    \qend)$
  }{
    $(\qreached, D)
      \leftarrow
        \VoxelizeFreeEdgeRecurse(\qbegin,
                                 \qend)
      \!\!\!\!\!\!\!\!\!\!\!\!\!\!\!\!\!\!\!\!\!\!\!\!
      $ \;
    $D \leftarrow D \cup \VoxelizeBackbone(\qreached)$ \;
    \Return{$(\qreached, D)$} \;
  }

  \tcp{Recursive implementation. $D$ is the voxelization}
  \tcp{from $\qbegin$ up to (but not including) $\qreached$}
  \myfunc{$\VoxelizeFreeEdgeRecurse(
    \qbegin,
    \qend)$
  }{
    $\Delta_{\text{voxel}} \leftarrow$ voxel distance from~\eqref{eqn:voxel-distance} \;
    \If(\tcp*[f]{Base case}){$\|\qend - \qbegin\| \leq \delta$
        \KwOr
        $\Delta_{\text{voxel}} \leq 1$
    }{
      \If(\tcp*[f]{collision-check}){$\qend$ is in $Q_{\text{free}}$}{
        \label{alg:voxelize-free-edge:collision-check}
        \Return{$(\qend, \VoxelizeBackbone(\qbegin))$} \;
      }
      \Return{$(\qbegin, \emptyset)$}
        \tcp*[r]{collision or invalid config}
    }

    $\qmid \leftarrow$ halfway between $\qbegin$ and $\qend$ \;
    $(\qreached, D)
      \leftarrow
        \VoxelizeFreeEdgeRecurse(
          \qbegin,
          \qmid
        )
      \!\!\!\!\!\!\!\!\!\!\!\!\!\!\!\!\!\!\!\!\!\!\!\!
      $ \;
    \If(\tcp*[f]{no collision, continue}){$\qreached = \qmid$}{
      $(\qreached, D_2)
        \leftarrow
          \VoxelizeFreeEdgeRecurse(
            \qmid,
            \qend
          )
        \!\!\!\!\!\!\!\!\!\!\!\!\!\!\!\!\!\!\!\!\!\!\!\!
        $ \;
      $D \leftarrow D \cup D_2$ \;
    }

    \Return{$(\qreached, D)$} \;
  }
\end{algorithm}

Our dynamic edge discretizer is implemented by the \VoxelizeFreeEdge algorithm (Alg.~\ref{alg:voxelize-free-edge}) which voxelizes a swept-volume motion of the robot backbone from $\qbegin$ to $\qend$ until detecting a collision.
The algorithm returns the voxelized swept-volume $D$ and the last collision-free configuration $\qreached$ before the detected collision.
In Alg.~\ref{alg:roadmap-ik}, \RoadmapIK voxelizes the edge from $\vec{q}_{\text{nn}}$ toward $\vec{q}_{\text{IK}}$ until detecting a collision.
This algorithm recursively subdivides the region until reaching a specified minimum configuration distance $\delta$ or the voxel distance between two robot shapes from~\eqref{eqn:voxel-distance} exceeds one.

In the Precompute phase, we do not have knowledge of the anatomical environment, and therefore during roadmap pre-voxelization, we replace $Q_{\text{free}}$ with $Q_{\text{valid}}$ on line~\ref{alg:voxelize-free-edge:collision-check} of Alg.~\ref{alg:voxelize-free-edge}.

For the evaluation of our dynamic edge discretizer, we use a discretization threshold $\delta$ of \SI{5e-4}{\N} in tension space, \SI{5e-3}{\mm} in retraction space, and \SI{5e-4}{\radian} in rotation space, separately evaluating distances in the three configuration subspaces.
These thresholds were chosen to be larger than the observed worst-case discretization from our dynamic edge strategy on \num{40 000} edges.
This discretization threshold is used as the constant step size in the discrete edge checker.
This facilitates a fair comparison, evaluating on discretizations that are similar in quality.

\section{Inverse Kinematics Algorithm}
\label{apdx:ik}

\begin{algorithm}[h]
  \caption{\NormalIK}
  \label{alg:normal-ik}

  \DontPrintSemicolon
  \SetArgSty{} 
  \SetDataSty{} 
  \SetKwInOut{Input}{input}
  \SetKwInOut{Output}{output}
  \SetKwComment{tcp}{\textbf{//} }{}
  \newcommand{\myAlgCommentSty}[1]{\color{blueishgreen}#1}
  \SetCommentSty{myAlgCommentSty}
  \SetFuncSty{normaltext}

  \SetKwFunction{IKsolve}{IK}
  \SetKwFunction{sampleUniform}{sample}
  \SetKwData{threshold}{threshold}

  \KwIn{\\
    \quad $\vec{g}$:       from $\mathds{R}^3$, goal tip position \\
    \quad $\vec{q}_{\text{current}}$:
                           from $Q_{\text{free}}$, current robot configuration \\
    \quad $k_{\text{restarts}}$: max number of random restarts for IK \\
    \quad \threshold:      tip error threshold \\
    \quad $\delta$:        discretization threshold distance \\
  }
  \KwOut{$q_{\text{IK}}$: collision-free IK solution for $g$}

  $\vec{q}_{\text{guess}} \leftarrow \vec{q}_{\text{current}}$ 
    \tcp*[r]{initial guess for IK}
  $Q_{\text{IK}} \leftarrow \emptyset$ 
    \tcp*[r]{IK solns. not within \threshold}
  \For{$i \leftarrow 0$ \KwTo $k_{\text{restarts}}$}{
    $\vec{q}_{\text{IK}} \leftarrow \IKsolve(\vec{q}_{\text{guess}}, \vec{g}, \threshold)$
      \tcp*[r]{using LM}
    $\vec{\hat{q}}_{\text{diff}}
      \leftarrow
        \frac{
          \vec{q}_{\text{guess}} - \vec{q}_{\text{IK}}
        }{
          \|\vec{q}_{\text{guess}} - \vec{q}_{\text{IK}}\|
        }$ 
      \tcp*[r]{unit vec. from $\vec{q}_{\text{IK}}$ to $\vec{q}_{\text{guess}}$}
    \While(\tcp*[f]{backstep until in $Q_{\text{free}}$})
          {$\vec{q}_{\text{IK}} \notin Q_{\text{free}}$}
    {
      $\vec{q}_{\text{IK}} \leftarrow \vec{q}_{\text{IK}} + \delta \vec{\hat{q}}_{\text{diff}}$ \;
    }
    \If{$\|\vec{T}_{\text{tip}}(\vec{q}_{\text{IK}}) - \vec{g}\| < \threshold$}{
      \Return{$\vec{q}_{\text{IK}}$}
        \tcp*[r]{success, exit early}
    } \Else {
      $Q_{\text{IK}} \leftarrow Q_{\text{IK}} \cup \{ \vec{q}_{\text{IK}} \}$ \;
      $\vec{q}_{\text{guess}} \leftarrow \sampleUniform(Q_\text{free})$ 
        \tcp*[r]{random restart}
    }
  }
  \Return{$
    \displaystyle \argmin_{\vec{q}_{\text{IK}} \in Q_{\text{IK}}}
    \|\vec{T}_{\text{tip}}(\vec{q}_{\text{IK}}) - \vec{g}\|
  $}
    \tcp*[r]{return closest}
\end{algorithm}

The \NormalIK algorithm (Alg.~\ref{alg:normal-ik}) is used by the RRTConnect, PRM, and LazyPRM planners in our comparison study in Section~\ref{sec:eval:comparison}.
\NormalIK represents a traditional IK approach for motion planning and has two key differences from \RoadmapIK (Alg.~\ref{alg:roadmap-ik}).
\begin{enumerate}
  \item
    \RoadmapIK leverages roadmap configurations that are nearby to the desired tip position $\vec{g}$, but \NormalIK uses only the current robot configuration $\vec{q}_{\text{current}}$ combined with $k_{\text{restarts}}$ random restarts.
  \item
    Both \RoadmapIK and \NormalIK utilize the LM algorithm as the underlying nonlinear optimizer which solves for configurations in $Q$ instead of $Q_{\text{free}}$.
    However, the two algorithms project their solution (which may be in collision) into $Q_{\text{free}}$ in different ways.
    \NormalIK steps backwards from the IK solution towards its initial guess until the solution is in $Q_{\text{free}}$ (note the initial guess comes from $Q_{\text{free}}$).
    \RoadmapIK instead does the opposite, effectively stepping toward the IK solution from the initial guess until detecting a collision, thus ensuring at least one collision-free edge from the roadmap to the returned numerical IK solution.
\end{enumerate}
\NormalIK approximates the solution to~\eqref{eqn:ik-problem}.
\RoadmapIK approximates~\eqref{eqn:ik-problem-mod} using the added constraint that there exists a linear collision-free edge from $\vec{q}_{\text{reached}}$ to the roadmap graph.

\NormalIK and \RoadmapIK use the same tip-error stopping threshold for the underlying LM optimizer.
The optimizer uses central differences for the numerical Jacobian with a finite difference distance four times smaller than the discrete sampler distance threshold.
Our method's \RoadmapIK uses \num{20} maximum iterations and initial LM damping factor $\lambda = 10$.
\NormalIK (used by PRM, LazyPRM, and RRTConnect) on the \num{25} goal set uses \num{200} maximum iterations and $\lambda = \num{4000}$, and on the \num{200} goal set uses \num{10} maximum iterations and $\lambda = \num{1000}$.

\section*{Acknowledgment}

The authors thank Dr. Chakravarthy Reddy for clinical insights and assistance with the anatomical environment used in evaluation.

\bibliographystyle{IEEEtran}
\bibliography{references}

\begin{thebibliography}{10}
\providecommand{\url}[1]{#1}
\csname url@samestyle\endcsname
\providecommand{\newblock}{\relax}
\providecommand{\bibinfo}[2]{#2}
\providecommand{\BIBentrySTDinterwordspacing}{\spaceskip=0pt\relax}
\providecommand{\BIBentryALTinterwordstretchfactor}{4}
\providecommand{\BIBentryALTinterwordspacing}{\spaceskip=\fontdimen2\font plus
\BIBentryALTinterwordstretchfactor\fontdimen3\font minus
  \fontdimen4\font\relax}
\providecommand{\BIBforeignlanguage}[2]{{%
\expandafter\ifx\csname l@#1\endcsname\relax
\typeout{** WARNING: IEEEtran.bst: No hyphenation pattern has been}%
\typeout{** loaded for the language `#1'. Using the pattern for}%
\typeout{** the default language instead.}%
\else
\language=\csname l@#1\endcsname
\fi
#2}}
\providecommand{\BIBdecl}{\relax}
\BIBdecl

\bibitem{Burgner-Kahrs2015_TRO}
\BIBentryALTinterwordspacing
J.~{Burgner-Kahrs}, D.~C. Rucker, and H.~Choset, ``Continuum robots for medical
  applications: A survey,'' \emph{IEEE Trans. Robot.}, vol.~31, no.~6, pp.
  1261--1280, Dec. 2015. [Online]. Available:
  \url{https://doi.org/10.1109/TRO.2015.2489500}
\BIBentrySTDinterwordspacing

\bibitem{Dupont2022_ProcIEEE}
\BIBentryALTinterwordspacing
P.~Dupont, N.~Simaan, H.~Choset, and C.~Rucker, ``Continuum robots for medical
  interventions,'' \emph{Proc. IEEE}, vol. 110, no.~7, pp. 847--870, Feb. 2022.
  [Online]. Available: \url{https://doi.org/10.1109/JPROC.2022.3141338}
\BIBentrySTDinterwordspacing

\bibitem{Kato2015_TM}
\BIBentryALTinterwordspacing
T.~Kato, I.~Okumura, S.-E. Song, A.~J. Golby, and N.~Hata, ``Tendon-driven
  continuum robot for endoscopic surgery: Preclinical development and
  validation of a tension propagation model,'' \emph{IEEE/ASME Trans.
  Mechatron.}, vol.~20, no.~5, pp. 2252--2263, Oct. 2015. [Online]. Available:
  \url{https://doi.org/10.1109/TMECH.2014.2372635}
\BIBentrySTDinterwordspacing

\bibitem{Kutzer2011_ICRA}
\BIBentryALTinterwordspacing
M.~D. Kutzer, S.~M. Segreti, C.~Y. Brown, M.~Armand, R.~H. Taylor, and S.~C.
  Mears, ``Design of a new cable-driven manipulator with a large open lumen:
  Preliminary applications in the minimally-invasive removal of osteolysis,''
  in \emph{Proc. 2011 {{IEEE}} Int. Conf. on Robot. and Autom. ({{ICRA}})},
  {Shanghai, China}, May 2011, pp. 2913--2920. [Online]. Available:
  \url{https://doi.org/10.1109/ICRA.2011.5980285}
\BIBentrySTDinterwordspacing

\bibitem{Nguyen2015_IROS}
\BIBentryALTinterwordspacing
T.-D. Nguyen and J.~{Burgner-Kahrs}, ``A tendon-driven continuum robot with
  extensible sections,'' in \emph{Proc. 2015 {{IEEE}}/{{RSJ}} Int. Conf. on
  Intell. Robots and Sys. ({{IROS}})}, {Hamburg, Germany}, Sep. 2015, pp.
  2130--2135. [Online]. Available:
  \url{https://doi.org/10.1109/IROS.2015.7353661}
\BIBentrySTDinterwordspacing

\bibitem{Oliver-Butler2019_TRO}
\BIBentryALTinterwordspacing
K.~{Oliver-Butler}, J.~Till, and C.~Rucker, ``Continuum robot stiffness under
  external loads and prescribed tendon displacements,'' \emph{IEEE Trans.
  Robot.}, vol.~35, no.~2, pp. 403--419, Apr. 2019. [Online]. Available:
  \url{https://doi.org/10.1109/TRO.2018.2885923}
\BIBentrySTDinterwordspacing

\bibitem{Rucker2011_TRO}
\BIBentryALTinterwordspacing
D.~C. Rucker and R.~J. Webster, III, ``Statics and dynamics of continuum robots
  with general tendon routing and external loading,'' \emph{IEEE Trans.
  Robot.}, vol.~27, no.~6, pp. 1033--1044, Dec. 2011. [Online]. Available:
  \url{https://doi.org/10.1109/TRO.2011.2160469}
\BIBentrySTDinterwordspacing

\bibitem{Huang2021_ISMR}
\BIBentryALTinterwordspacing
Y.~Huang, M.~Bentley, T.~Hermans, and A.~Kuntz, ``Toward learning
  context-dependent tasks from demonstration for tendon-driven surgical
  robots,'' in \emph{Proc. 2021 Int. Symp. on Med. Robot. ({{ISMR}})},
  {Atlanta, GA, USA}, Nov. 2021. [Online]. Available:
  \url{https://doi.org/10.1109/ISMR48346.2021.9661534}
\BIBentrySTDinterwordspacing

\bibitem{Starke2017_IROS}
\BIBentryALTinterwordspacing
J.~Starke, E.~Amanov, M.~T. Chikhaoui, and J.~{Burgner-Kahrs}, ``On the merits
  of helical tendon routing in continuum robots,'' in \emph{Proc. 2017
  {{IEEE}}/{{RSJ}} Int. Conf. on Intell. Robots and Sys. ({{IROS}})},
  {Vancouver, BC, Canada}, Sep. 2017, pp. 6470--6476. [Online]. Available:
  \url{https://doi.org/10.1109/IROS.2017.8206554}
\BIBentrySTDinterwordspacing

\bibitem{Lynch2017_book}
K.~M. Lynch and F.~C. Park, \emph{Modern Robotics: Mechanics, Planning, and
  Control}.\hskip 1em plus 0.5em minus 0.4em\relax {Cambridge, England}:
  {Cambridge Univ. Press}, 2017.

\bibitem{Torres2014_ICRA}
\BIBentryALTinterwordspacing
L.~G. Torres, C.~Baykal, and R.~Alterovitz, ``Interactive-rate motion planning
  for concentric tube robots,'' in \emph{Proc. 2014 {{IEEE}} Int. Conf. on
  Robot. and Autom. ({{ICRA}})}, {Hong Kong, China}, May 2014, pp. 1915--1921.
  [Online]. Available: \url{https://doi.org/10.1109/ICRA.2014.6907112}
\BIBentrySTDinterwordspacing

\bibitem{Torres2015_ICRA}
\BIBentryALTinterwordspacing
L.~G. Torres, A.~Kuntz, H.~B. Gilbert, P.~J. Swaney, R.~J. Hendrick, R.~J.
  Webster, III, and R.~Alterovitz, ``A motion planning approach to automatic
  obstacle avoidance during concentric tube robot teleoperation,'' in
  \emph{Proc. 2015 {{IEEE}} Int. Conf. on Robot. and Autom. ({{ICRA}})},
  {Seattle, WA, USA}, May 2015, pp. 2361--2367. [Online]. Available:
  \url{https://doi.org/10.1109/ICRA.2015.7139513}
\BIBentrySTDinterwordspacing

\bibitem{Peters2018_SE}
\BIBentryALTinterwordspacing
B.~S. Peters, P.~R. Armijo, C.~Krause, S.~A. Choudhury, and D.~Oleynikov,
  ``Review of emerging surgical robotic technology,'' \emph{Surg. Endosc.},
  vol.~32, no.~4, pp. 1636--1655, Feb. 2018. [Online]. Available:
  \url{https://doi.org/10.1007/s00464-018-6079-2}
\BIBentrySTDinterwordspacing

\bibitem{Zhang2021_JAIT}
\BIBentryALTinterwordspacing
J.~Zhang, K.~Lu, J.~Yuan, J.~Di, and G.~He, ``Kinematics modeling and motion
  planning of tendon driven continuum manipulators,'' \emph{J. Artif. Intell.
  Technol.}, vol.~1, no.~1, pp. 28--36, 2021. [Online]. Available:
  \url{https://doi.org/10.37965/jait.2020.0041}
\BIBentrySTDinterwordspacing

\bibitem{Neppalli2009_AR}
\BIBentryALTinterwordspacing
S.~Neppalli, M.~A. Csencsits, B.~A. Jones, and I.~D. Walker, ``Closed-form
  inverse kinematics for continuum manipulators,'' \emph{Adv. Robot.}, vol.~23,
  no.~15, pp. 2077--2091, 2009. [Online]. Available:
  \url{https://doi.org/10.1163/016918609X12529299964101}
\BIBentrySTDinterwordspacing

\bibitem{Xu2009_JMR}
\BIBentryALTinterwordspacing
K.~Xu and N.~Simaan, ``Analytic formulation for kinematics, statics, and shape
  restoration of multibackbone continuum robots via elliptic integrals,''
  \emph{J. Mech. Robot.}, vol.~2, no.~1, Feb. 2010. [Online]. Available:
  \url{https://doi.org/10.1115/1.4000519}
\BIBentrySTDinterwordspacing

\bibitem{Lai2022_RAL}
\BIBentryALTinterwordspacing
J.~Lai, B.~Lu, Q.~Zhao, and H.~Chu, ``Constrained motion planning of a
  cable-driven soft robot with compressible curvature modeling,'' \emph{IEEE
  Robot. Autom. Lett.}, vol.~7, no.~2, pp. 4813--4820, Apr. 2022. [Online].
  Available: \url{https://doi.org/10.1109/LRA.2022.3152318}
\BIBentrySTDinterwordspacing

\bibitem{Ashwin2019_AMMS}
\BIBentryALTinterwordspacing
K.~P. Ashwin and A.~Ghosal, ``Profile estimation of a cable-driven continuum
  robot with general cable routing,'' in \emph{Proc. 15th {{IFToMM}} World
  Congr. Mech. Mach. Sci.}, ser. Mechanisms and {{Machine Science}}, T.~Uhl,
  Ed., vol.~73.\hskip 1em plus 0.5em minus 0.4em\relax {Krak\'ow, Poland}:
  {Springer International Publishing}, Jun. 2019, pp. 1879--1888. [Online].
  Available: \url{https://doi.org/10.1007/978-3-030-20131-9_186}
\BIBentrySTDinterwordspacing

\bibitem{Mahapatra2022_AAMMS}
\BIBentryALTinterwordspacing
S.~K. Mahapatra, K.~P. Ashwin, and A.~Ghosal, ``Modelling of~cable-driven
  continuum robots with~general cable routing: A~comparison,'' in \emph{Proc.
  {{IFToMM Asian MMS}} 2021}, ser. Mechanisms and Machine Science, N.~V. Khang,
  N.~Q. Hoang, and M.~Ceccarelli, Eds., vol. 113.\hskip 1em plus 0.5em minus
  0.4em\relax {Hanoi, Vietnam}: {Springer, Cham}, Jan. 2022, pp. 345--353.
  [Online]. Available: \url{https://doi.org/10.1007/978-3-030-91892-7_32}
\BIBentrySTDinterwordspacing

\bibitem{LaValle1998_Report}
\BIBentryALTinterwordspacing
S.~M. LaValle, ``Rapidly-exploring random trees : A new tool for path
  planning,'' {Comput. Sci. Dept., Iowa State Univ.}, {Ames, IA, USA}, Tech.
  Rep.~11, Oct. 1998. [Online]. Available:
  \url{http://msl.cs.uiuc.edu/~lavalle/papers/Lav98c.pdf}
\BIBentrySTDinterwordspacing

\bibitem{Kavraki1996_TRA}
\BIBentryALTinterwordspacing
L.~E. Kavraki, P.~{\v S}vestka, J.-C. Latombe, and M.~H. Overmars,
  ``Probabilistic roadmaps for path planning in high-dimensional configuration
  spaces,'' \emph{IEEE Trans. Robot. Autom.}, vol.~12, no.~4, pp. 566--580,
  Aug. 1996. [Online]. Available: \url{https://doi.org/10.1109/70.508439}
\BIBentrySTDinterwordspacing

\bibitem{Kuffner2000_ICRA}
\BIBentryALTinterwordspacing
J.~J. Kuffner, Jr. and S.~M. LaValle, ``{{RRT-connect}}: An efficient approach
  to single-query path planning,'' in \emph{Proc. 2000 {{IEEE}} Int. Conf. on
  Robot. and Autom. ({{ICRA}})}, vol.~2, {San Francisco, CA, USA}, Apr. 2000,
  pp. 995--1001. [Online]. Available:
  \url{https://doi.org/10.1109/ROBOT.2000.844730}
\BIBentrySTDinterwordspacing

\bibitem{Bohlin2000_ICRA}
\BIBentryALTinterwordspacing
R.~Bohlin and L.~E. Kavraki, ``Path planning using lazy {{PRM}},'' in
  \emph{Proc. 2000 {{IEEE}} Int. Conf. on Robot. and Autom. ({{ICRA}})},
  vol.~1, {San Francisco, CA, USA}, Apr. 2000, pp. 521--528. [Online].
  Available: \url{https://doi.org/10.1109/ROBOT.2000.844107}
\BIBentrySTDinterwordspacing

\bibitem{Karaman2011_IJRR}
\BIBentryALTinterwordspacing
S.~Karaman and E.~Frazzoli, ``Sampling-based algorithms for optimal motion
  planning,'' \emph{Int. J. Rob. Res.}, vol.~30, no.~7, pp. 846--894, Jun.
  2022. [Online]. Available: \url{https://doi.org/10.1177/0278364911406761}
\BIBentrySTDinterwordspacing

\bibitem{Hauser2015_ICRA}
\BIBentryALTinterwordspacing
K.~Hauser, ``Lazy collision checking in asymptotically-optimal motion
  planning,'' in \emph{Proc. 2015 {{IEEE}} Int. Conf. on Robot. and Autom.
  ({{ICRA}})}, {Seattle, WA, USA}, May 2015, pp. 2951--2957. [Online].
  Available: \url{https://doi.org/10.1109/ICRA.2015.7139603}
\BIBentrySTDinterwordspacing

\bibitem{Deng2019_RoboSoft}
\BIBentryALTinterwordspacing
J.~Deng, B.~H. Meng, I.~Kanj, and I.~S. Godage, ``Near-optimal smooth path
  planning for multisection continuum arms,'' in \emph{Proc. 2019 2nd {{IEEE}}
  Int. Conf. on Soft Robot. ({{RoboSoft}})}, {Seoul, South Korea}, Apr. 2019,
  pp. 416--421. [Online]. Available:
  \url{https://doi.org/10.1109/ROBOSOFT.2019.8722778}
\BIBentrySTDinterwordspacing

\bibitem{Meng2021_ICRA}
\BIBentryALTinterwordspacing
B.~H. Meng, I.~S. Godage, and I.~Kanj, ``Smooth path planning for continuum
  arms,'' in \emph{Proc. 2021 {{IEEE}} Int. Conf. on Robot. and Autom.
  ({{ICRA}})}, {Xi'an, China}, May 2021, pp. 7809--7814. [Online]. Available:
  \url{https://doi.org/10.1109/ICRA48506.2021.9560982}
\BIBentrySTDinterwordspacing

\bibitem{Sun2015_TRO}
\BIBentryALTinterwordspacing
W.~Sun, S.~Patil, and R.~Alterovitz, ``High-frequency replanning under
  uncertainty using parallel sampling-based motion planning,'' \emph{IEEE
  Trans. Robot.}, vol.~31, no.~1, pp. 104--116, Feb. 2015. [Online]. Available:
  \url{https://doi.org/10.1109/TRO.2014.2380273}
\BIBentrySTDinterwordspacing

\bibitem{Patil2014_TRO}
\BIBentryALTinterwordspacing
S.~Patil, J.~Burgner, R.~J. Webster, III, and R.~Alterovitz, ``Needle steering
  in 3-{{D}} via rapid replanning,'' \emph{IEEE Trans. Robot.}, vol.~30, no.~4,
  pp. 853--864, Aug. 2014. [Online]. Available:
  \url{https://doi.org/10.1109/TRO.2014.2307633}
\BIBentrySTDinterwordspacing

\bibitem{Li2018_MBEC}
\BIBentryALTinterwordspacing
P.~Li, Z.~Yang, and S.~Jiang, ``Needle-tissue interactive mechanism and
  steering control in image-guided robot-assisted minimally invasive surgery: A
  review,'' \emph{Med. Biol. Eng. Comput.}, vol.~56, no.~6, pp. 931--949, Jun.
  2018. [Online]. Available: \url{https://doi.org/10.1007/s11517-018-1825-0}
\BIBentrySTDinterwordspacing

\bibitem{Leibrandt2017_RAM}
\BIBentryALTinterwordspacing
K.~Leibrandt, C.~Bergeles, and G.-Z. Yang, ``Concentric tube robots: Rapid,
  stable path-planning and guidance for surgical use,'' \emph{IEEE Robot.
  Autom. Mag.}, vol.~24, no.~2, pp. 42--53, Jun. 2017. [Online]. Available:
  \url{https://doi.org/10.1109/MRA.2017.2680546}
\BIBentrySTDinterwordspacing

\bibitem{Nocedal2006_Book_ch10}
\BIBentryALTinterwordspacing
J.~Nocedal and S.~Wright, ``Least-squares problems,'' in \emph{Numerical
  Optimization}.\hskip 1em plus 0.5em minus 0.4em\relax {New York, NY}:
  {Springer}, 2006, pp. 245--269. [Online]. Available:
  \url{https://doi.org/10.1007/978-0-387-40065-5}
\BIBentrySTDinterwordspacing

\bibitem{Amanatides1987_EG}
J.~Amanatides and A.~Woo, ``A fast voxel traversal algorithm for ray tracing,''
  in \emph{Proc. 8th Eur. Comput. Graphics Conf. Exhib. ({{Eurographics}})},
  {Amsterdam, the Netherlands}, Aug. 1987.

\bibitem{Meagher1982_CGIP}
\BIBentryALTinterwordspacing
D.~Meagher, ``Geometric modeling using octree encoding,'' \emph{Comput.
  Graphics and Image Process.}, vol.~19, no.~2, pp. 129--147, Jun. 1982.
  [Online]. Available: \url{https://doi.org/10.1016/0146-664X(82)90104-6}
\BIBentrySTDinterwordspacing

\bibitem{Voelker2017_Report}
\BIBentryALTinterwordspacing
A.~R. Voelker, J.~Gosmann, and T.~C. Stewart, ``Efficiently sampling vectors
  and coordinates from the n-sphere and n-ball,'' {Centre for Theoretical
  Neuroscience, Columbia Univ.}, {New York, NY, USA}, Tech. Rep., Jan. 2017.
  [Online]. Available: \url{https://doi.org/10.13140/RG.2.2.15829.01767/1}
\BIBentrySTDinterwordspacing

\bibitem{Shunsuke2012_JRM}
\BIBentryALTinterwordspacing
S.~Toritani, R.~L.~A. Shauri, K.~Nonami, and D.~Fujiwara, ``Numerical solution
  using nonlinear least-squares method for inverse kinematics calculation of
  redundant manipulators,'' \emph{J. Robot. Mech.}, vol.~24, no.~2, pp.
  363--371, Apr. 2012. [Online]. Available:
  \url{https://doi.org/10.20965/jrm.2012.p0363}
\BIBentrySTDinterwordspacing

\bibitem{Pinter2019_CMPB}
\BIBentryALTinterwordspacing
C.~Pinter, A.~Lasso, and G.~Fichtinger, ``Polymorph segmentation representation
  for medical image computing,'' \emph{Comput. Methods and Programs in
  Biomed.}, vol. 171, pp. 19--26, Apr. 2019. [Online]. Available:
  \url{https://doi.org/10.1016/j.cmpb.2019.02.011}
\BIBentrySTDinterwordspacing

\bibitem{Pan2012_ICRA}
\BIBentryALTinterwordspacing
J.~Pan, S.~Chitta, and D.~Manocha, ``{{FCL}}: A general purpose library for
  collision and proximity queries,'' in \emph{Proc. 2012 {{IEEE}} Int. Conf. on
  Robot. and Autom. ({{ICRA}})}, {St. Paul, MN, USA}, May 2012, pp. 3859--3866.
  [Online]. Available: \url{https://doi.org/10.1109/ICRA.2012.6225337}
\BIBentrySTDinterwordspacing

\bibitem{Sucan2012_RAM}
\BIBentryALTinterwordspacing
I.~A. Sucan, M.~Moll, and L.~E. Kavraki, ``The open motion planning library,''
  \emph{IEEE Robot. Autom. Mag.}, vol.~19, no.~4, pp. 72--82, Dec. 2012.
  [Online]. Available: \url{https://doi.org/10.1109/MRA.2012.2205651}
\BIBentrySTDinterwordspacing

\bibitem{Franco2014_IJCSI}
\BIBentryALTinterwordspacing
E.~Franco, F.~Madera, and F.~{Moo-Mena}, ``Self-collision detection in tubular
  objects approximated by spheres,'' \emph{Int. J. Comput. Sci. Issues},
  vol.~11, no.~5, pp. 14--21, Sep. 2014. [Online]. Available:
  \url{http://www.ijcsi.org/papers/IJCSI-11-5-1-14-21.pdf}
\BIBentrySTDinterwordspacing

\bibitem{Kuntz2019_IROS}
\BIBentryALTinterwordspacing
A.~Kuntz, M.~Fu, and R.~Alterovitz, ``Planning high-quality motions for
  concentric tube robots in point clouds via parallel sampling and
  optimization,'' in \emph{Proc. 2019 {{IEEE}}/{{RSJ}} Int. Conf. on Intell.
  Robots and Sys. ({{IROS}})}, {Macau, China}, Nov. 2019, pp. 2205--2212.
  [Online]. Available: \url{https://doi.org/10.1109/IROS40897.2019.8968172}
\BIBentrySTDinterwordspacing

\bibitem{Kuntz2020_ISRR}
\BIBentryALTinterwordspacing
A.~Kuntz, C.~Bowen, and R.~Alterovitz, ``Fast anytime motion planning in point
  clouds by interleaving sampling and interior point optimization,'' in
  \emph{Robot. Res.}, ser. Springer Proc. in Adv. Robot. ({{SPAR}}), N.~M.
  Amato, G.~Hager, S.~Thomas, and M.~{Torres-Torriti}, Eds., vol.~10.\hskip 1em
  plus 0.5em minus 0.4em\relax {Springer, Cham}, Jan. 2020, pp. 929--945.
  [Online]. Available: \url{https://doi.org/10.1007/978-3-030-28619-4_63}
\BIBentrySTDinterwordspacing

\bibitem{Ahnert2011_AIP}
\BIBentryALTinterwordspacing
K.~Ahnert and M.~Mulansky, ``Odeint--solving ordinary differential equations in
  {{C}}++,'' in \emph{{{AIP}} Conference Proceedings}, vol. 1389, {Halkidiki,
  Greece}, Sep. 2011, pp. 1586--1589. [Online]. Available:
  \url{https://doi.org/10.1063/1.3637934}
\BIBentrySTDinterwordspacing

\bibitem{Lourakis2004_code}
\BIBentryALTinterwordspacing
M.~Lourakis, ``Levmar: {{Levenberg-Marquardt}} nonlinear least squares
  algorithms in {{C}}/{{C}}++,'' Jul. 2004. [Online]. Available:
  \url{http://www.ics.forth.gr/~lourakis/levmar/}
\BIBentrySTDinterwordspacing

\end{thebibliography}

\begin{IEEEbiography}[\authfig{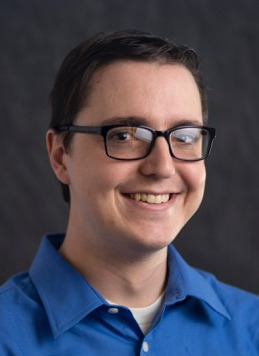}]{Michael Bentley}
  received his B.S. degree in mathematics and his B.A. degree in physics with a minor in computer science from the University of Utah, Salt Lake City, UT, USA, in 2012.
  He is currently a Ph.D. candidate for a degree in computing in the robotics track in the School of Computing at the University of Utah.
  His research interests include robot modeling for medical applications, safe motion planning, continuum robots, and software reproducibility.
\end{IEEEbiography}

\begin{IEEEbiography}[\authfig{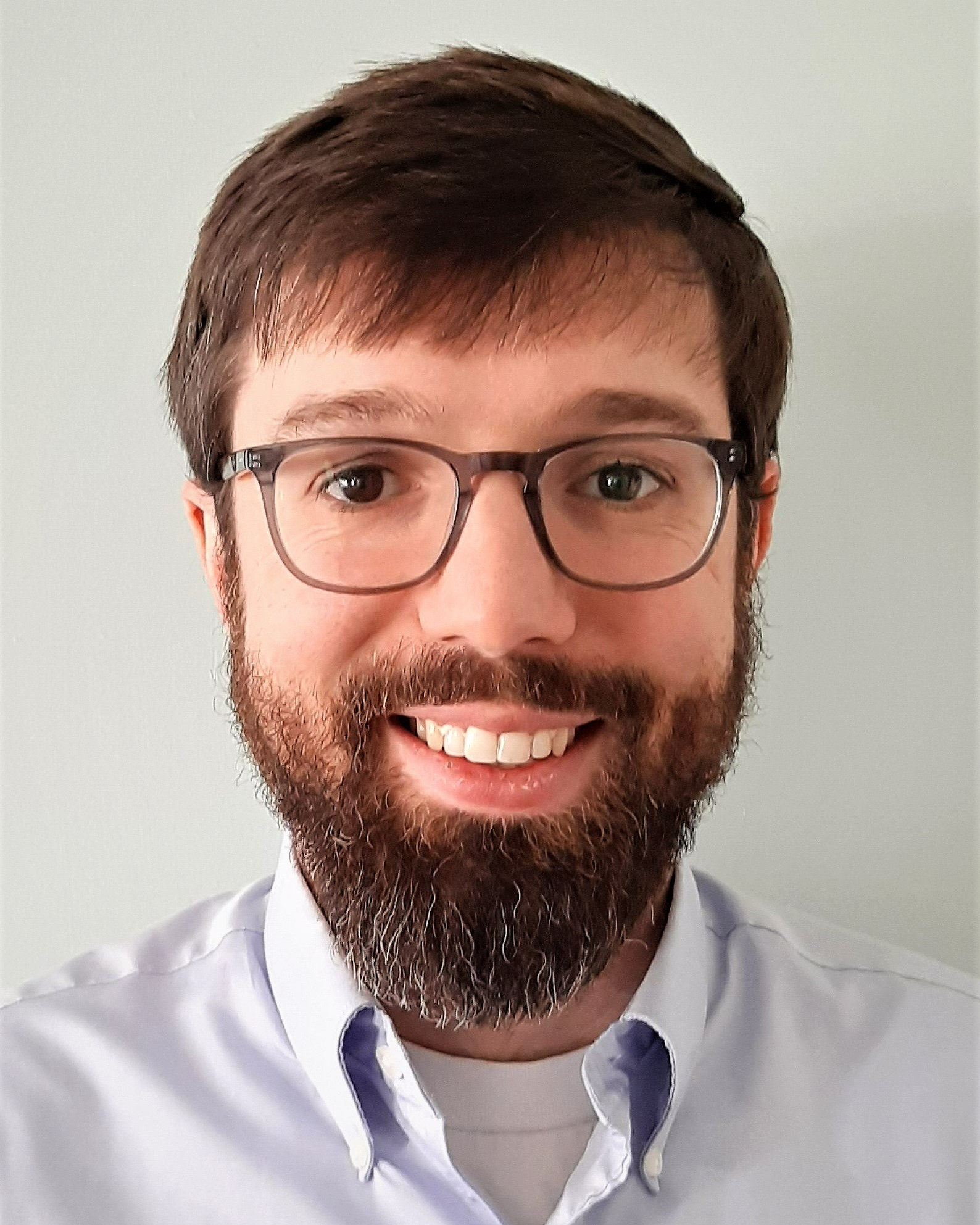}]{Caleb Rucker}
  (M'13, SM'22)
  received his B.S. degree in engineering mechanics and mathematics from Lipscomb University, Nashville, TN, USA, in 2006, and his Ph.D. degree in mechanical engineering from Vanderbilt University, Nashville, TN, USA, in 2010.

  From 2011 to 2013, he was a postdoctoral fellow in Biomedical Engineering at Vanderbilt University.
  He is now an Associate Professor in mechanical engineering at the University of Tennessee, Knoxville, TN, USA, where he directs the Robotics, Engineering, and Continuum Mechanics in Healthcare Laboratory (REACH Lab).
  He received the NSF CAREER award in 2017.
\end{IEEEbiography}

\begin{IEEEbiography}[\authfig{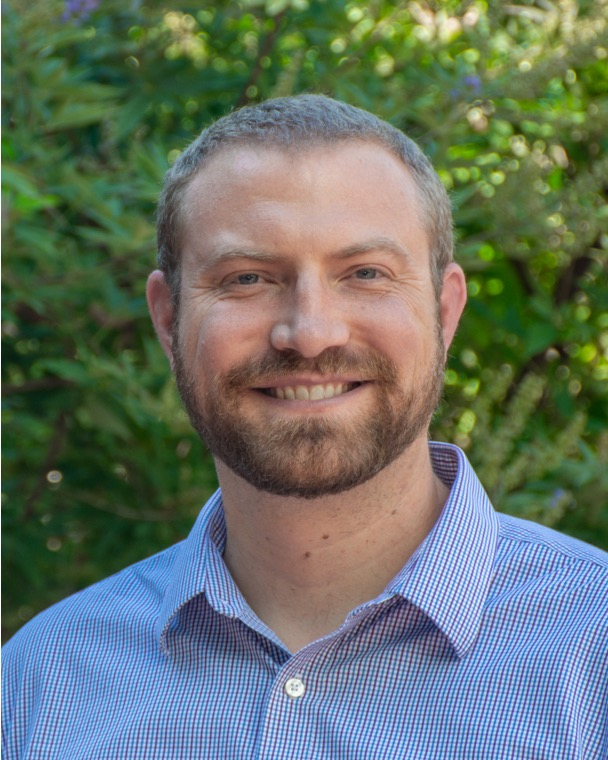}]{Alan Kuntz} (M'19)
  received his B.S. degree in computer science with Honors from the University of New Mexico, Albuquerque, NM, USA, in 2014, his M.S. degree in computer science from the University of North Carolina at Chapel Hill, Chapel Hill, NC, USA, in 2016, and his Ph.D. degree in computer science from the University of North Carolina at Chapel Hill in 2019.

  In 2020, he joined the faculty of the School of Computing and the Robotics Center at the University of Utah, Salt Lake City, UT, USA, as an Assistant Professor, where he leads the artificial intelligence and robotics in medicine lab.
  Prior to joining the University of Utah, he was a postdoctoral scholar at the Vanderbilt Institute for Surgery and Engineering and the Department of Mechanical Engineering at Vanderbilt University, Nashville, TN, USA.
  His research focuses on healthcare applications of artificial intelligence, design optimization, and robot motion planning.
\end{IEEEbiography}

\EOD

\end{document}